\documentclass[twoside,11pt]{article}

\usepackage{jmlr2e}

\usepackage{natbib,mathrsfs,amsmath,bm,amssymb,algorithm,algorithmic}
\usepackage{smile}
\usepackage{jmlr2e}
\usepackage{lastpage}
\usepackage{threeparttable}
\usepackage[protrusion=false,expansion=true]{microtype}
\usepackage{Tabbing}
\usepackage{chngcntr}
\usepackage{graphicx} 
\usepackage{epstopdf}
\usepackage{comment}
\usepackage{undertilde}
\usepackage[usenames,dvipsnames]{xcolor}
\hypersetup{
  pdftex,
  pdffitwindow=true,
  pdfstartview={FitH},
  pdfnewwindow=true,
  colorlinks,
  linktocpage=true,
  citecolor=blue
}

\usepackage[width=\textwidth ,font={small},labelfont={color=blue,bf,up},labelsep=endash, format=plain, labelsep=period,
textfont=it, font=singlespacing]{caption}

\usepackage{bookmark}

\newcommand{\ra}[1]{\renewcommand{\arraystretch}{#1}}

\usepackage{lastpage}
\jmlrheading{22}{2021}{1-\pageref{LastPage}}{9/19; Revised
1/21}{1/21}{19-769}{Botao Hao, Boxiang Wang, Pengyuan Wang, Jingfei Zhang, Jian Yang, Will Wei Sun}

\ShortHeadings{Sparse Tensor Additive Regression}{Hao, Wang, Wang, Zhang, Yang and Sun}

\firstpageno{1}

\begin{document}

\title{ Sparse Tensor Additive Regression}

\author{\name Botao Hao \email haobotao000@gmail.com \\
	\addr
	Deepmind\\
	5 New Street, London, UK
	\AND
	\name Boxiang Wang \email boxiang-wang@uiowa.edu \\
	\addr Department of Statistics and Actuarial Science\\
	The University of Iowa\\
    Iowa City, IA 52242, USA
	\AND
	\name Pengyuan Wang \email pengyuan@uga.edu \\
	\addr Department of Marketing\\
	University of Georgia\\
   Athens, GA 30602, USA
	\AND
	\name Jingfei Zhang \email ezhang@bus.miami.edu \\
	\addr Department of Management Science\\
	University of Miami\\
   Coral Gables, FL 33146, USA
	\AND
	\name Jian Yang \email  jianyang@verizonmedia.com \\
	\addr Yahoo Research \\
Verizon Media \\
   Sunnyvale, CA 94089, USA
	\AND
	\name Will Wei Sun \email sun244@purdue.edu \\
	   \addr Krannert School of Management\\
	    Purdue University\\
	   West Lafayette, IN 47907, USA
	   }

\editor{Francis Bach}

\maketitle

\begin{abstract}
Tensors are becoming prevalent in modern applications such as medical imaging and digital marketing. In this paper, we propose a sparse tensor additive regression (STAR) that models a scalar response as a flexible nonparametric function of tensor covariates. The proposed model effectively exploits the sparse and low-rank structures in the tensor additive regression. We formulate the parameter estimation as a non-convex optimization problem, and propose an efficient penalized alternating minimization algorithm. We establish a non-asymptotic error bound for the estimator obtained from each iteration of the proposed algorithm, which reveals an interplay between the optimization error and the statistical rate of convergence. We demonstrate the efficacy of STAR through extensive comparative simulation studies, and an application to the click-through-rate prediction in online advertising.
\end{abstract}


\section{Introduction}\label{sec:intro}
Tensor data have recently become popular in a wide range of applications such as medical imaging \citep{zhou2013, li2017parsimonious, sun2017store}, digital marketing \citep{zhe2016distributed, sun2017provable}, video processing \citep{guo2012tensor}, and social network analysis \citep{park2009pairwise, hoff2015multilinear}, among many others. In such applications, a fundamental statistical tool is \textit{tensor regression}, a modern high-dimensional regression method that relates a scalar response to tensor covariates. For example, in neuroimaging analysis, an important objective is to predict clinical outcomes using subjects' brain imaging data. This can be formulated as a tensor regression problem by treating the clinical outcomes as the response and the brain images as the tensor covariates. Another example is in the study of how advertisement placement affect users' clicking behavior in online advertising. This again can be formulated as a tensor regression problem by treating the daily overall click-through rate (CTR) as the response and the tensor that summarizes the impressions (i.e., view counts) of different advertisements on different devices (e.g., phone, computer, etc.) as the covariate.
In Section~\ref{sec:real}, we consider such an online advertising application.

\begin{figure}[!htb]
  \begin{minipage}[b]{\linewidth}
    \centering
    \includegraphics[width=1\textwidth]{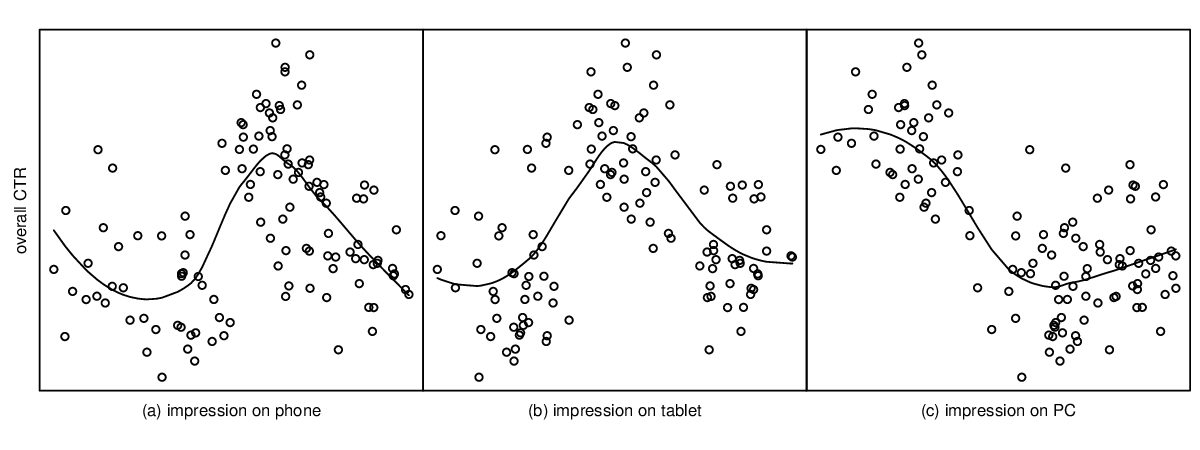}
    \caption{\small The overall click-through rate v.s. the impression of a certain advertisement that is delivered on phone (left plot), tablet (middle plot), and PC (right plot), respectively. The black solid curves are the fitted locally weighted scatter-plot smoother (LOESS) curves.}
    \label{fig:nonlinear}
  \end{minipage}
\end{figure}

Denote $y_i$ as a scalar response and $\cX_i \in \mathbb R^{p_1 \times p_2 \ldots \times p_m}$ as an $m$-way tensor covariate, $i=1,2,\ldots, n$. A general tensor regression model can be formulated as
\begin{equation*}
y_i = \cT^*(\cX_i) + \epsilon_i, \ i=1,2,\ldots, n,
\end{equation*}
where $\cT^*(\cdot): \mathbb R^{p_1 \times p_2 \ldots \times p_m}\rightarrow \mathbb R$ is an unknown regression function, $\{\epsilon_i\}_{i=1}^n$ are scalar observation noises.  
Many existing methods assumed a linear relationship between the response and the tensor covariates by considering $\cT^*(\cX_i) = \langle \cB, \cX_i\rangle$ for some low-rank tensor coefficient $\cB$ \citep{zhou2013, rabusseau2016,yu2016, guhaniyogi2017bayesian, raskutti2019convex}. 
In spite of its simplicity, the linear assumption could be restrictive and difficult to satisfy in real applications. 
Consider the online advertising data in Section~\ref{sec:real} as an example. 
Figure~\ref{fig:nonlinear} shows the marginal relationship between the overall CTR and the impressions of an advertisement delivered on phone, tablet, and PC, respectively. 
It is clear that the relationship between the response (i.e., the overall CTR) and the covariate (i.e., impressions across three devices) departs notably from the linearity assumption. 
A few work considered more flexible tensor regressions by treating $\cT^*(\cdot)$ as a nonparametric function \citep{suzuki2016minimax,kanagawa2016gaussian}. In particular, \cite{suzuki2016minimax} proposed a general nonlinear model where the true function $\cT^*(\cdot)$ is consisted of components from a reproducing kernel Hilbert space, and used an alternating minimization estimation procedure; \cite{kanagawa2016gaussian} considered a Bayesian approach that employed a Gaussian process prior in learning the nonparametric function $\cT^*(\cdot)$ on the reproducing kernel Hilbert space. One serious limitation of both work is that they assume that the tensor covariates are \emph{exact low-rank}. This assumption is difficult to satisfy in practice, as most tensor covariates are not exact low-rank. When the tensor covariates are not exact low-rank, the performance of these two methods deteriorates dramatically; see Section~\ref{sec:full_rank} for more details. In addition, the Gaussian process approach is computationally very expensive, which severely limits its application in problems with high-dimensional tensor covariates.

In this paper, we develop a flexible and computationally feasible tensor regression framework, which accommodates the nonlinear relationship between the response and the tensor covariate, and is highly interpretable. Specifically, for an $m$-way tensor covariate $\cX_i \in \mathbb{R}^{p_1 \times \ldots \times p_m}$, we consider a sparse tensor additive regression (STAR) model with
\begin{equation}\label{eqn:STAR}
\cT^*(\cX_i) = \sum_{j_1=1}^{p_1}\cdots\sum_{j_m=1}^{p_m}f^*_{j_1\ldots j_m}([\cX_i]_{j_1\ldots j_m}), 
\end{equation}
where $[\cX_i]_{j_1\ldots j_m}$ denotes the $(j_1,\ldots, j_m)$-th element of $\cX_i$, and $f_{j_1\ldots j_m}^*(\cdot)$ is a nonparametric additive component belonging to some smooth function class. Approximating the additive component $f_{j_1\ldots j_m}^*(\cdot)$ using spline series expansions, $\cT^*(\cX_i)$ can be simplified to have a compact tensor representation of spline coefficients. To reduce the number of parameters and increase computational efficiency, we assume that the corresponding high-dimensional coefficient tensors have low-rank and group sparsity structures. Both low-rankness and sparsity are commonly used dimension reduction tools in recent tensor models \citep{li2017parsimonious, sun2017provable, sun2017store, hao2018sparse, zhang2019cross, zhang2019optimal}. Besides effectively reducing computational cost, the group sparsity structure also significantly improves the model interpretability. For instance, in the online advertising example, when the daily overall CTR is regressed on the impressions of different advertisements on different devices, the group sparsity enables our STAR model to select effective advertisement and device combinations. Such a type of advertisement selection is important for managerial decision making and has been an active research area \citep{choi2010using, xu2016lift}. 
To efficiently estimate the model, we formulate the parameter estimation as a non-convex optimization and propose a penalized alternating minimization algorithm. By fully exploiting the low-rankness and group sparsity structures as well as developing an efficient algorithm, our STAR model may run faster than the tensor linear regression in some experiments. For example, in the online advertising application, our STAR model can reduce the CTR prediction error by 50\% while using 10\% computational time of the linear or nonlinear tensor regression benchmark models. See Section~\ref{sec:real} for more details.

Besides methodological contributions, we also obtain some strong theoretical results for our proposed method. In particular, we first establish a general theory for penalized alternating minimization in the context of tensor additive model. To the best of our knowledge, this is the first statistical-versus-optimization guarantee for the penalized alternating minimization. Previous work mostly focus on either the EM-type update \citep{wang2014optimal, BWY15, hao2017simultaneous}, or the truncation-based update \citep{sun2017provable}. Those techniques are not directly applicable to our scenario; see Section \ref{subsec:opt_stat} for detailed explanations. Next, we derive a non-asymptotic error bound for the estimator from each iteration, which demonstrates the improvement of the estimation error in each update. Finally, we apply this general theory to our STAR estimator with B-spline basis and the group-lasso penalty, and show that the estimation error in the $(t+1)$-th iteration satisfies
$$
\cE^{(t+1)}\leq\underbrace{\rho^{t+1}\cE^{(0)}}_{\text{optimization error}}+ \underbrace{\frac{C_1}{1-\rho}n^{-\tfrac{2\kappa-1}{2\kappa+1}}\log (pd_n)}_{\text{statistical error}},
$$
where $0<\rho\leq 1/2$ is a contraction parameter, $\kappa$ is the smoothness parameter of the function class, $p = \max\{p_1,\ldots, p_m\}$, and $d_n$ is the number of spline series. The above error bound reveals an interesting interplay between the optimization error and the statistical error. The optimization error decays geometrically with the iteration number $t$, while the statistical error remains the same as $t$ grows. When the tensor covariate is of order one (i.e., a vector covariate), our problem reduces to the vector nonparametric additive model. 
In that case, our statistical error matches with that from the vector nonparametric additive model in \cite{huang2010variable}.

\subsection{Other related work}

The problem we consider in our work is fundamentally different from those in tensor decomposition and tensor response regression. As a result, the technical tools involved and the theoretical results are quite different. 

Tensor decomposition \citep{chi2012on, anandkumar2014tensor, yuan2016on, sun2017provable} is an unsupervised learning method that aims to find the best low-rank approximation of a single tensor. In comparison, our STAR model is a supervised learning method that seeks to capture the nonlinear relationship between the response and the tensor covariate. Although the low-rank structure of the tensor coefficient is also employed in our estimation, our objective and the technical tools involved are entirely different from the typical tensor decomposition problem. 
Additionally, one fundamental difference is that our model works with multiple tensor samples, while tensor decomposition works only with a single tensor. As a result, our error bound is a function of the sample size, which is different from that in tensor decomposition.

Another line of related work considers tensor response regression, where the response is a tensor and the covariates are scalars \citep{ZhuHT2009intrinsic, li2017parsimonious, sun2017store}. 
These work also utilized the low-rank and/or sparse structures of the coefficient tensors for dimension reduction. 
However, tensors are treated as the \emph{response} in tensor response regression, whereas they are treated as a \emph{covariates} in our approach. These are two very different types of models, motivated by different applications. 
The tensor response regression aims to study the change of the tensor (e.g., the brain image) as the covariate (e.g., disease status) varies. 
However, the tensor regression model focuses on understanding the change of a scalar outcome (e.g., the overall CTR) with the tensor covariates. As a result, technical tools used for theoretical analysis are also largely different.

\subsection{Notations and structure}

Throughout this article, we denote scalars by lower case characters such as $x$, vectors by lower-case bold characters such as $\bx$, matrices by upper-case bold characters such as $\bX$ and tensors by upper-case script characters such as $\cX$. Given a vector $\bx\in\mathbb R^p$ and a set of indices $T\subset \{1,\ldots, p\}$, we define $\bx_T$ such that $x_{T_j}=x_j$ if $j\in T$ and $x_{T_j}=0$, otherwise. For a square matrix $\bA$, we denote $\sigma_{\min}(\bA)$ and $\sigma_{\max}(\bA)$ as its minimum and maximum eigenvalues, respectively.  For any function $f$ on $[a, b]$, we define its $\ell_2(P)$ norm by $\|f(x)\|_2 = \sqrt{\int_{a}^bf^2(x)dP(x)}$. Suppose $\cX,\cY\in\mathbb R^{p_1\times p_2\times \cdots \times p_m}$ are $m$-way tensors. We define tensor inner product $\langle\cX,\cY \rangle = \sum_{j_1,\ldots,j_m} \cX_{j_1\ldots j_m}\cY_{j_1\ldots j_m}$. The tensor Frobenius norm is defined as $\|\cX\|_F = \sqrt{\sum_{j_1=1}^{p_1}\cdots\sum_{j_m=1}^{p_m}\cX_{j_1\ldots j_m}^2}$. The notation $a\lesssim b$ implies $a\leq C_1b$ for some constant $C_1>0$. For any two sequences $\{a_n\}_{n=1}^\infty, \{b_n\}_{n=1}^\infty$, we write $a_n = \cO(b_n)$ if there exists some positive constant $C_2$ and sufficiently large $n$ such that $a_n\leq C_2b_n$. We also write $a_n\asymp b_n$ if there exist constants $C_3, C_4>0$ such that $C_3a_n\leq b_n \leq C_4a_n$ for all $n\geq 1$. 

The rest of the article is organized as follows. Section~\ref{sec:model} introduces our sparse tensor additive regression model. Section~\ref{sec:est} develops an efficient penalized alternating minimization algorithm for model estimation. 
Section~\ref{sec:theory} investigates its theoretical properties, followed by simulation studies in Section~\ref{sec:simu} and a real online advertising application in Section~\ref{sec:real}. The appendix collects all technical proofs.

\section{Sparse Tensor Additive Model}\label{sec:model}
Given i.i.d. samples $\{y_i, \cX_i\}_{i=1}^n$, our sparse tensor additive model assumes
\begin{equation}\label{eqn:model}
y_i = \cT^*(\cX_i) + \epsilon_i =\sum_{j_1=1}^{p_1}\cdots\sum_{j_m=1}^{p_m}f^*_{j_1\ldots j_m}([\cX_i]_{j_1\ldots j_m}) + \epsilon_i, \ i=1,\ldots n,
\end{equation}
where $f_{j_1\ldots j_m}^*(\cdot)$ is the nonparametric additive function belonging to some smooth function class ${\cal H}$, and $\{\epsilon_i\}_{i=1}^n$ are i.i.d. observation noises. 

Our STAR model utilizes spline series expansion \citep{huang2010variable,fan2011nonparametric} to approximate each individual nonparametric additive component. Let $\cS_n$ be the space of polynomial splines and $\{\psi_{h}(x)\}_{h=1}^{d_n}$ be a normalized basis for $\cS_n$, where $d_n$ is the number of spline series and $\sup_x|\psi_{h}(x)|\leq 1$. 
It is known that for any $f_n\in\cS_n$, there always exists some coefficients $\{\beta_{h}^*\}_{h=1}^{d_n}$ such that
$f_{n}(x) = \sum_{h=1}^{d_n}\beta_{h}^*\psi_{h}(x)$. 
In addition, under suitable smoothness assumptions (see Lemma \ref{lemma:spline_appro}), each nonparametric additive component $f_{j_1\ldots j_m}^*(\cdot)$  can be well approximated by functions in $\cS_n$.  Applying the above approximation to each individual component, the regression function $\cT^*(\cX_i) $ in \eqref{eqn:model} can be approximated by
\begin{equation}\label{eqn:addtive}
\cT^*(\cX_i)  \approx \sum_{j_1=1}^{p_1}\cdots\sum_{j_m=1}^{p_m}\sum_{h=1}^{d_n}\beta^*_{j_1\ldots j_mh}\psi_{j_1\ldots j_mh}([\cX_i]_{j_1\ldots j_m}).
\end{equation}
The expression in $(\ref{eqn:addtive})$ has a compact tensor representation. Define $\cF_h(\cX)\in\mathbb{R}^{p_1\times\ldots\times p_m}$ such that $[\cF_h(\cX)]_{j_1\ldots j_m} = \psi_{j_1\ldots j_mh}([\cX]_{j_1\ldots j_m})$, and $\cB_h^*\in\mathbb{R}^{p_1\times\ldots\times p_m}$ such that $[\cB_h^*]_{j_1\ldots j_m} = \beta_{j_1\ldots j_mh}^*$ for $h \in[d_n]$, where $[k]$ denotes $\{1,\ldots, k\}$ for an integer $k \ge 1$. Consequently, we can write
\begin{equation}\label{eqn:F_n}
\sum_{j_1=1}^{p_1}\cdots\sum_{j_m=1}^{p_m}\sum_{h=1}^{d_n}\beta^*_{j_1\ldots j_mh}\psi_{j_1\ldots j_mh}([\cX_i]_{j_1\ldots j_m}) = \sum_{h=1}^{d_n} \Big \langle\cB^*_h, \cF_h(\cX_i)\Big\rangle.
\end{equation}
Therefore, the parameter estimation of the nonparametric additive model \eqref{eqn:model} reduces to the estimation of unknown tensor coefficients $\cB_1^*,\ldots,\cB_{d_n}^*$. The coefficients $\cB_1^*,\ldots,\cB_{d_n}^*$ include a total number of $\cO(d_n\Pi_{j=1}^mp_j)$ free parameters, which could be much larger than the sample size $n$. In such ultrahigh-dimensional scenario, it is important to employ dimension reduction tools. A common tensor dimension reduction tool is the low-rank assumption \citep{chi2012on, anandkumar2014tensor, yuan2016on, sun2017provable}. Similarly, we assume each coefficient tensor $\cB_1^*,\ldots,\cB_{d_n}^*$ satisfies the CP low-rank decomposition \citep{kolda2009}:
\begin{eqnarray}\label{eqn:cp_dec}
\cB_h^* = \sum_{r=1}^R\bbeta_{1hr}^*\circ\cdots\circ\bbeta_{mhr}^*, \ h=1,\ldots,d_n,
\end{eqnarray}
where $\circ$ is the vector outer product, $\bbeta_{1hr}^*\in \mathbb R^{p_1}, \ldots, \bbeta_{mhr}^*\in \mathbb R^{p_m}$, and $R\ll \min\{p_1,\ldots, p_m\}$ is the CP-rank. This formulation reduces the effective number of the parameters from $\cO(d_n\Pi_{j=1}^mp_j)$ to $\cO(d_nR\sum_{j=1}^mp_j)$, and hence greatly improves computational efficiency. Under this formulation, our model can be written as
\begin{eqnarray}\label{eqn:multilinear}
    \cT^*(\cX_i) \approx \sum_{h=1}^{d_n}  \Big\langle \cF_h(\cX_i), \sum_{r=1}^R\bbeta_{1hr}^*\circ\cdots\circ\bbeta_{mhr}^*\Big\rangle.
\end{eqnarray}

\begin{remark}
Our model in $(\ref{eqn:multilinear})$ can be viewed as a generalization of several existing work. When $\psi_{j_1\ldots j_mh}(\cdot)$ in \eqref{eqn:F_n} is an identity basis function ($\psi_{j_1\ldots j_mh}([\cX]_{j_1\ldots j_m}) = [\cX]_{j_1\ldots j_m}$) with only one basis ($d_n=1$), \eqref{eqn:multilinear} reduces to the bilinear form \citep{li2010dimension, hung2012matrix} for a matrix covariate ($m=2$), and the multilinear form for linear tensor regression \citep{zhou2013, hoff2015multilinear, yu2016, rabusseau2016, sun2017store, guhaniyogi2017bayesian, raskutti2019convex} for a tensor covariate ($m\ge3$).
\end{remark}

\begin{figure}[h]
  \begin{minipage}[b]{\linewidth}
    \centering
    \includegraphics[width=0.6\textwidth]{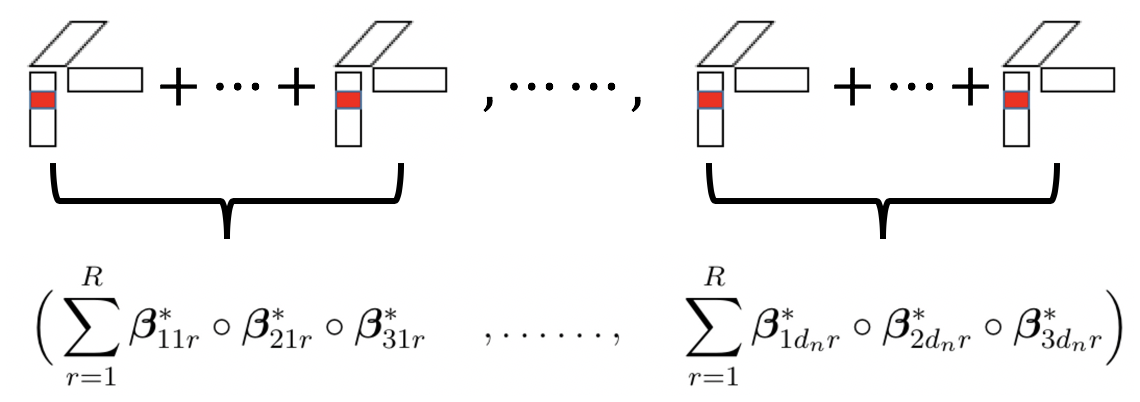}
    \caption{\small An illustration of the low-rank and group-sparsity structures in a collection of three-way tensor coefficients $(\cB_1^*,\ldots, \cB_{d_n}^*)$. If one or more of the coefficients at the colored locations are non-zero, the cardinality of $\cS_1$ increases by one. }
    \label{fig:sparsity}
  \end{minipage}
\end{figure}

In addition to the CP low-rank structure on the tensor coefficients, we further impose a group-type sparsity constraint on the components $\bbeta_{khr}^*$. This group sparsity structure not only further reduces the effective parameter size, but also improves the model interpretability, as it enables the variable selection of components in the tensor covariate. Recall that in \eqref{eqn:multilinear} we have $\bbeta_{khr}^* = (\beta_{khr1}^*, \ldots, \beta_{khrp_k}^*)^{\top}$ for $k\in[m], h\in[d_n], r\in[R]$. We define our group sparsity constraint as
\begin{equation}\label{def:beta}
\Big| \underbrace{\Big\{j\in[p_k]\Big| \sum_{h=1}^{{d_n}}\sum_{r=1}^{R}\beta_{khrj}^{*2}\neq 0\Big\}}_{\cS_k}\Big|=s_k\ll p_k, \ \text{for} \ k\in[m].
\end{equation}
where $|\cS_k|$ refers to the cardinality of the set $\cS_k$. Figure \ref{fig:sparsity} provides an illustration of the low-rank \eqref{eqn:cp_dec} and group-sparse \eqref{def:beta} coefficients when the order of the tensor is $m=3$. When $m=1$, our model with the group sparsity constraint reduces to the vector sparse additive model \citep{Ravikumar2009, meier2009high, huang2010variable}.
\begin{remark}
 Consider the tensor order as 2. The model in \eqref{eqn:multilinear} reduces to
\begin{eqnarray*}
    \cT^*(\cX_i) \approx \sum_{h=1}^{d_n}  \Big\langle \cF_h(\cX_i), \sum_{r=1}^R\bbeta_{1hr}^*\bbeta_{2hr}^{*\top}\Big\rangle\in\mathbb R^{p_1\times p_2}.
\end{eqnarray*}
For instance, suppose $\sum_{h=1}^{{d_n}}\sum_{r=1}^{R}\beta_{1hr1}^{*2} = 0$.
This implies the first row of RHS is all 0 and thus encourages variable selection in $ \cT^*(\cX_i)$ correspondingly.
\end{remark}

\section{Estimation}\label{sec:est}
In this section, we describe our approach to estimate the parameters in our STAR model via a penalized empirical risk minimization which simultaneously satisfies the low-rankness and encourages the sparsity of decomposed components. In particular, we consider
\begin{equation}\label{def:ERM}
\min_{\bbeta_{1hr},\ldots,\bbeta_{mhr}} \underbrace{\frac{1}{n}\sum_{i=1}^n\Big(y_i-\sum_{h=1}^{d_n}\Big\langle\sum_{r=1}^R\bbeta_{1hr}\circ\cdots\circ\bbeta_{mhr}, \cF_h(\cX_i)\Big\rangle\Big)^2}_{\cL(\bbeta_{1hr}, \ldots, \bbeta_{mhr})}+ \cP(\bbeta_{1hr}, \ldots, \bbeta_{mhr}),
\end{equation}
where $\cL(\bbeta_{1hr}, \ldots, \bbeta_{mhr})$ is the empirical risk function, in which the low-rankness is guaranteed due to the CP decomposition, and $\cP(\cdot)$ is a penalty term that encourages sparsity. 
To enforce the sparsity as defined in \eqref{def:beta}, we consider the group lasso penalty \citep{yuan2006model}, i.e., 
\begin{equation}\label{eqn:group_lasso}
\cP(\bbeta_{1hr}, \ldots, \bbeta_{mhr}) = \sum_{k=1}^m\Big(\lambda_{kn}\sum_{j=1}^{p_k}\sqrt{\sum_{h=1}^{{d_n}}\sum_{r=1}^{R}\beta_{khrj}^2}\Big),
\end{equation} 
where $\{\lambda_{kn}\}_{k=1}^m$ are tuning parameters. It is worth mentioning that our algorithm and theoretical analysis can accommodate a general class of decomposable penalties (see Condition \ref{con:decomposable} for details), which includes lasso, ridge, fused lasso, and group lasso as special cases. 

For a general tensor covariate ($m>1$), the optimization problem in \eqref{def:ERM} is a non-convex optimization. This is fundamentally different from the vector sparse additive model \citep{Ravikumar2009, huang2010variable} whose optimization is convex. The non-convexity in \eqref{def:ERM} brings significant challenges in both model estimation and theoretical development. The key idea of our estimation procedure is to explore the bi-convex structure of the empirical risk function $\cL(\bbeta_{1hr}, \ldots, \bbeta_{mhr})$ since it is convex in one argument while fixing all the other parameters. This motivates us to rewrite the empirical risk function into a bi-convex representation, which in turn facilitates the introduction of an efficient alternating minimization algorithm. 

Denote $\vartheta_{krj} = (\beta_{k1rj}, \beta_{k2rj}, \ldots, \beta_{kd_nrj})^{\top}\in \mathbb R^{d_n\times 1}, \vartheta_{kj} = (\vartheta_{k1j}^{\top}, \ldots, \vartheta_{kRj}^{\top})^{\top}\in\mathbb R^{Rd_n\times 1}$ for $k\in[m]$, $j\in[p_k]$, and $\bb_k = (\vartheta_{k1}^{\top}, \ldots, \vartheta_{kp_k}^{\top})^{\top}$. We also define the operator $\prod_{k\in[m]}^{\circ}\ba_k=\ba_1\circ\cdots\circ\ba_m$.  Remind that $\cF_h(\cX)\in\mathbb{R}^{p_1\times\ldots\times p_m}$ with $[\cF_h(\cX)]_{j_1\ldots j_m} = \psi_{j_1\ldots j_mh}([\cX]_{j_1\ldots j_m})$, see $(\ref{eqn:F_n})$.   We use $[\cF_{h}^k(\cX_i)]_{j}$ to refer to the $m-1$ way tensor when we fix the index along the $k$-th way of $\cF_{h}(\cX_i)$ as $j$, e.g., $[\cF_{h}^1(\cX_i)]_{j} \in \mathbb{R}^{p_2\times\ldots\times p_m}$ . Define
  \begin{eqnarray*}
  F_{irj}^k = \Big(\big\langle\prod_{u\in[m]\setminus k}^{\circ}\bbeta_{u1r}, [\cF_1^k(\cX_i)]_{j}\big\rangle, \ldots, \big\langle\prod_{u\in[m]\setminus k}^{\circ}\bbeta_{ud_nr}, [\cF_{d_n}^k(\cX_i)]_{j}\big\rangle\Big)^{\top},
  \end{eqnarray*}
and denote $F_{ij}^k = (F_{i1j}^{k\top}, \ldots, F_{iRj}^{k\top})^{\top} \in \mathbb R^{Rd_n\times 1}$. In addition, we denote $\bF_j^k = (F_{1j}^k, \ldots, F_{nj}^k)^{\top}$,  $\bF^k=(\bF_1^k, \ldots, \bF_{p_k}^k)$, and $\by = (y_1, \ldots,y_n)^{\top}$. Thus, when other parameters are fixed, minimizing the empirical risk function \eqref{def:ERM} with respect to $\bb_k$ is equivalent to minimizing
\begin{equation}\label{eqn:rewrite_risk}
\cL\big(\bb_1, \cdots, \bb_m\big)=\frac{1}{n}\|\by-\bF^k\bb_k\|_2^2.
\end{equation}
Note that the expression of \eqref{eqn:rewrite_risk} holds for any $k\in[m]$ with proper definitions on $\bF^k$ and $\bb_k$.
\begin{remark}
Intuitively, $\vartheta_{kj}$ summarizes all the colored coefficients in Figure \ref{fig:sparsity} and $\bb_k$ summarizes all the coefficients along $k$-th mode. By this definition, we can more clearly describe the effect of group sparsity: when $\vartheta_{kj}$ is a zero vector, the $j$-th variable in $k$-th mode is irrelevant.
\end{remark}

Based on this reformulation, we are ready to introduce the alternating minimization algorithm that solves \eqref{def:ERM} by alternatively updating $\bb_1,\cdots, \bb_m$. A desirable property of our algorithm is that updating $\bb_k$ given others can be solved efficiently via the group-wise coordinate descent based on the back-fitting algorithm \citep{Ravikumar2009}. The detailed algorithm is summarized in Algorithm \ref{alg:main}. With a little abuse of notations, we redefine the penalty term $\cP(\bb_k^{(t)}) = \sum_{j=1}^{p_k}\sqrt{\sum_{h=1}^{d_n}\sum_{r=1}^R(\beta_{khrj}^{(t)})^2}$. 

\begin{algorithm}
\small
\caption{Penalized Alternating Minimization for Solving \eqref{def:ERM}}
\begin{algorithmic}[1]
  \STATE \textbf{Input:} $\{y_i\}_{i=1}^n$, $\{\cX_i\}_{i=1}^n$, initialization $\{\bb_1^{(0)}, \ldots, \bb_m^{(0)}\}$, the set of penalization parameters $\{\lambda_{1n}, \ldots, \lambda_{mn}\}$, rank $R$, iteration $t=0$, stopping error $\epsilon = 10^{-5}$.
  \STATE Repeat $t= t+1$ and run penalized alternating minimization.
  \STATE \quad For $k=1$ to $m$
  \begin{eqnarray}\label{eqn:update_b}
    \bb_k^{(t+1)} = \argmin_{\bb_k}\cL(\bb_1^{(t)}, \ldots, \bb_m^{(t)}) + \lambda_{kn}\cP(\bb_k^{(t)}),
  \end{eqnarray}
    \quad \quad where $\cL$ is defined in \eqref{eqn:rewrite_risk}. 
  \STATE \quad End for.
    \STATE Until $\max_{k}\|\bb_k^{(t+1)}-\bb_k^{(t)}\|_2\leq \epsilon$ , and let $t=T^*$.
    \STATE \textbf{Output:} the estimate of each component, $\{\bb_1^{(T^*)}, \ldots, \bb_m^{(T^*)}\}$.
  \end{algorithmic}\label{alg:main}
\end{algorithm}

In our implementation, we use ridge regression to initialize Algorithm \ref{alg:main}, and set tuning parameters $\lambda = \lambda_{kn}$ for $k=1,\ldots,m$ for simplicity.  
When solving Problem~\eqref{eqn:update_b}, we use the \textit{warm start} and \textit{active set} to accelerate the algorithm. The basic idea of the two tricks is illustrated as follows: for each $t$, we use the solution $\bb_k^{(t-1)}$ as the initial value to update $\bb_k^{(t)}$, i.e., the warm start; when computing $\bb_k^{(t)}$, we may only consider an active set of $k$ such that $\bb_k^{(t-1)}$ is nonzero. Those two tricks have shown great successes in the implementations of the coordinate-descent-type algorithms such as the R package \texttt{glmnet} \citep{glmnet}.  The overall computational complexity of Algorithm \ref{alg:main} is $O(Tmkn(Rp)^m)$, where $T$ is the number of iterations for alternating minimization, $m$ is the order the tensor, $k$ is the number of iteration for group-wise coordinate descent to solve Problem~\eqref{eqn:update_b}, $n$ is the sample size, $R$ is the tensor rank and $p$ is the maximum dimension of each tensor mode.

\section{Theory}\label{sec:theory}
In this section, we first establish a general theory for the penalized alternating minimization in the context of the tensor additive model. Several sufficient conditions are proposed to guarantee the optimization error and statistical error. Then, we apply our theory to the STAR estimator with B-spline basis functions and the group-lasso penalty. To ease the presentation, we consider a three-way tensor covariate (i.e, $m=3$) in our theoretical development, while its generalization to an $m$-way tensor is straightforward. 


\subsection{A general contract property}\label{subsec:opt_stat}  

To bound the optimization error and statistical error of the proposed estimator, we introduce three sufficient conditions: a Lipschitz-gradient condition, a sparse strongly convex condition, and a generic statistical error condition. For the sake of brevity, we only present conditions for the update of $\bb_1$ in the main paper, and defer similar conditions for $\bb_2, \bb_3$ to Section \ref{subsec:add_con} in the appendix.

\begin{figure}[h]
  \begin{minipage}[b]{\linewidth}
    \centering
    \includegraphics[width=0.5\textwidth]{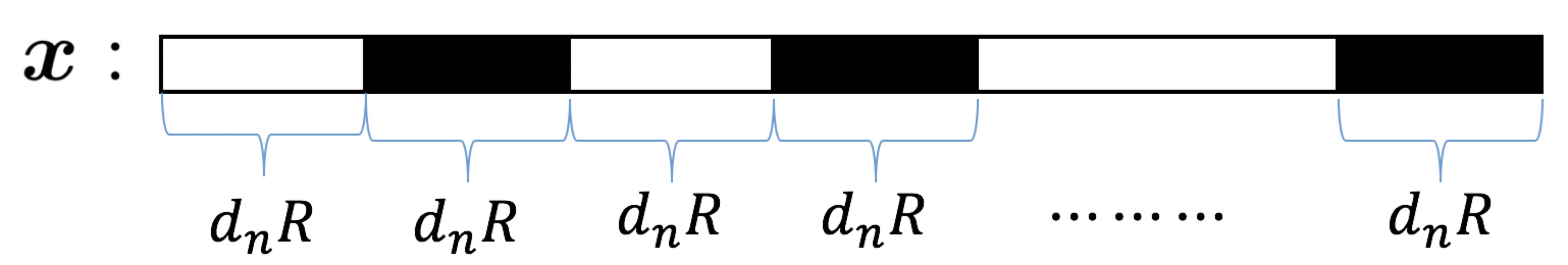}
    \caption{\small An illustration of the group sparse vector. A segment is colored if it contains at least one non-zero element, and a segment is uncolored if all of its elements are zero.}
    \label{fig:group_vector}
  \end{minipage}
\end{figure}

For each vector $\bx\in\mathbb R^{pd_nR\times 1}$, we divide it into $p$ equal-length segments as in Figure \ref{fig:group_vector}. 
A segment is colored if it contains at least one non-zero element, and a segment is uncolored if all of its elements are zero. 
We let $w(\bx)$ be the indices of colored segments in $\bx$ and $E_s$ be the set of all $(pd_nR)$-dimensional vectors with less than $C_0s$ colored segments, for some constant $C_0$.
Mathematically, for a vector $\bx\in \mathbb R^{pd_nR\times 1}$, denote $w(\bx) :=\{j\in[p]|\sum_{h=1}^{d_nR}x_{(j-1)d_nR+h}^2\neq 0\}$ and $E_s := \{\bx\in \mathbb R^{pd_nR\times 1} | |w(\bx)|\leq C_0 s\}$. 

Define a sparse ball $\cB_{\alpha, s}(\bb^*):=\{\bb\in\mathbb R^{pd_nR}:\|\bb-\bb^*\|_2\leq \alpha, \bb\in E_{s}\}$ for a given constant radius $\alpha$. Moreover, the noisy gradient function and noiseless gradient function of empirical risk function $\cL$ defined in \eqref{def:ERM} of order-3 with respect to $\bb_1$ can be written as
\begin{eqnarray}\label{def:gradient}
    && \nabla_1\cL(\bb_1, \bb_2, \bb_3)  = \frac{2}{n}\bF^{1\top}\Big(\bF^1\bb_1-\by\Big)\\
    && \nabla_1\tilde{\cL}(\bb_1, \bb_2, \bb_3)  = \frac{2}{n}\bF^{1\top}\Big(\bF^1\bb_1-\bF^{1*}\bb_1^*\Big),\nonumber
\end{eqnarray}
  where  $F_{irj}^{1*} = \big(\langle\bbeta_{21r}^*\circ\bbeta_{31r}^*, [\cF_1^1(\cX_i)]_{j}\rangle, \ldots, \langle\bbeta_{2d_nr}^*\circ\bbeta_{3d_nr}^*, [\cF_{d_n}^1(\cX_i)]_{j}\rangle\big)^{\top}.$
  
\begin{condition}[(Lipschitz-Gradient)]\label{con:gradient_stability} For $\bb_2\in\cB_{\alpha_2,s_2}(\bb_2^*), \bb_3\in\cB_{\alpha_3, s_3}(\bb_3^*)$, the noiseless gradient function $\nabla_1\tilde{\cL}(\bb_1^*, \cdot, \bb_3^*)$ satisfies $\mu_{2n}$-Lipschitz-gradient condition, and $\nabla_1\tilde{\cL}(\bb_1^*, \bb_2, \cdot)$ satisfies $\mu_{3n}$-Lipschitz-gradient condition with high probability. That is,
  \begin{eqnarray*}
  &&\big\langle\nabla_1\tilde{\cL}(\bb_1^*, \bb_2, \bb_3^*)-\nabla_1\tilde{\cL}(\bb_1^*, \bb_2^*, \bb_3^*), \bb_1-\bb_1^* \big\rangle\leq \mu_{2n}\big\|\bb_1-\bb_1^*\big\|_2\big\|\bb_2-\bb_2^*\big\|_2\\
  &&\big\langle\nabla_1\tilde{\cL}(\bb_1^*, \bb_2, \bb_3)-\nabla_1\tilde{\cL}(\bb_1^*, \bb_2, \bb_3^*), \bb_1-\bb_1^*\big\rangle\leq \mu_{3n}\big\|\bb_1-\bb_1^*\big\|_2\big\|\bb_3-\bb_3^*\big\|_2,
  \end{eqnarray*}
with probability at least $1-\delta_1$ for any $0<\delta_1<1$. Here, $\mu_{2n},\mu_{3n}$ may depend on $\delta_1$.
\end{condition}
\begin{remark}
Condition \ref{con:gradient_stability} defines a variant of Lipschitz continuity for $\nabla_1\tilde{\cL}(\bb_1^*, \cdot, \bb_3^*)$ and $\nabla_1\tilde{\cL}(\bb_1^*, \bb_2, \cdot)$. Note that the gradient is always taken with respect to the first argument of $\cL(\cdot, \cdot,\cdot)$ and the Lipschitz continuity is with respect to the second or the third argument. Analogous Lipschitz-gradient conditions were also considered in \cite{BWY15, hao2017simultaneous} for the \emph{population-level $Q$-function} in the EM-type update. 
\end{remark}

Next condition characterizes the curvature of noisy gradient function in a sparse ball. It states that when the second and the third argument are fixed, $\cL(\cdot, \cdot, \cdot)$ is strongly convex with parameter $\gamma_{1n}$ with high probability. As shown later in Section~\ref{sec:thm_STAR}, this condition holds for a broad family of basis functions.
\begin{condition}[(Sparse-Strong-Convexity)]\label{con:strongly_convex}
  For any $\bb_2\in\cB_{\alpha_2,s_2}(\bb_2^*), \bb_3\in\cB_{\alpha_3,s_3}(\bb_3^*)$, the loss function $\cL(\cdot,\cdot,\cdot)$ is sparse strongly convex in its first argument, namely
  \begin{eqnarray*}
    \cL(\bb_1^*, \bb_2, \bb_3)-\cL(\bb_1, \bb_2, \bb_3)-\langle\nabla_1 \cL\big(\bb_1^*, \bb_2, \bb_3), \bb_1^*-\bb_1\rangle\geq \frac{\gamma_{1n}}{2}\|\bb_1-\bb_1^*\|_2^2,
  \end{eqnarray*}
  with probability at least $1-\delta_2$ for any $0<\delta_2<1$. Here, $\gamma_{1n}>0$ is the strongly convex parameter and may depend on $\delta_2$.
\end{condition}
Next we present the definition for dual norm, which is a key measure for statistical error condition. More details on the dual norm are referred to \cite{negahban2012}. 
\begin{definition}[Dual norm]\label{def:dual_norm}
  For a given inner product $\langle\cdot, \cdot\rangle$, the dual norm of $\cP$ is given by
  \begin{equation*}
  \cP^*(\bv) :=\sup_{\bu\in\mathbb R^p\setminus \{0\}}\frac{\langle\bu,\bv\rangle}{\cP(\bu)}.
  \end{equation*}
\end{definition}
As a concrete example, the dual of $\ell_1$-norm is $\ell_{\infty}$-norm while the dual of $\ell_2$-norm is itself. Suppose $\bv$ is a $p$-dimensional vector and the index set $\{1,2,\ldots, p\}$ is partitioned into $N_{\cG}$ disjoint groups, namely $\cG = \{G_1,\ldots, G_{N_{\cG}}\}$. The group norm for $\bv$ is defined as $\cP(\bv) = \sum_{t=1}^{N_{\cG}}\|\bv_{G_t}\|_2$. According to Definition \eqref{def:dual_norm}, the dual of $\cP(\bv)$ is defined as $\cP^*(\bv) = \max_{t}\|\bv_{G_t}\|_2$. For simplicity, we write $\|\cdot\|_{\cP^*}=\cP^*(\cdot)$. 

The generic statistical error (SE) condition guarantees that the distance between noisy gradient and noiseless gradient under $\cP^*$-norm is bounded.
\begin{condition}[(Statistical-Error)]\label{con:stat_error}
  For any $\bb_2\in \cB_{\alpha_2,s_2}(\bb_2^*)$, $\bb_3\in\cB_{\alpha_3,s_3}(\bb_3^*)$, we have with probability at least $1-\delta_3$,
  \begin{equation*}
  \big\|\nabla_1\cL(\bb_1^*, \bb_2, \bb_3)-\nabla_1\tilde{\cL}(\bb_1^*, \bb_2, \bb_3)\big\|_{\cP^*}\leq \varepsilon_1.
  \end{equation*}
\end{condition}
\begin{remark}
Here, $\varepsilon_1$ is only a generic quantity and its explicit form will be derived for a specific loss function in Section \ref{sec:thm_STAR}. 
\end{remark}

Next we introduce two conditions for the penalization parameter (Condition \ref{con:regularization}) and penalty (Condition \ref{con:decomposable}). To illustrate Condition \ref{con:regularization}, we first introduce an quantity called \textit{support space compatibility constant} to measure the intrinsic dimensionality of $\cS_1$ defined in \eqref{def:beta} with respect to penalty $\cP$. Specifically, it is defined as 
\begin{eqnarray}\label{def:support_constant}
  \Phi(\cS_1):=\sup_{\bb\in \cS_1\setminus \{0\}}\frac{\cP(\bb)}{\|\bb\|_2},
\end{eqnarray}
which is a variant of subspace compatibility constant originally proposed by \cite{negahban2012} and \cite{Wain2014}. If $\cP(\bb)$ is chosen as a group lasso penalty, we have $\Phi(\cS')=\sqrt{|\cS'|}$, where $\cS'$ is the index set of active groups. Similar definitions of $\Phi(\cS_2), \Phi(\cS_3)$ can be made accordingly.
\begin{condition}[(Penalization Parameter)]\label{con:regularization} 
   We consider an iterative turning procedure where tuning parameters in $(\ref{eqn:update_b})$ are allowed to change with iteration. In particular, we assume tuning parameters $\{\lambda_{1n}^{(t)}, \lambda_{2n}^{(t)}, \lambda_{3n}^{(t)}\}$ satisfy 
   \begin{equation*}
   \begin{split}
   &\lambda_{1n}^{(t)}= 4\varepsilon_1+(\mu_{2n}\|\bb_2^{(t)}-\bb_2^*\|_2+\mu_{3n}\|\bb_3^{(t)}-\bb_3^*\|_2)/\Phi(\cS_1)\\
   &\lambda_{2n}^{(t)}= 4\varepsilon_2+(\mu'_{1n}\|\bb_1^{(t)}-\bb_1^*\|_2+\mu^{'}_{3n}\|\bb_3^{(t)}-\bb_3^*\|_2)/\Phi(\cS_2)\\
   &\lambda_{3n}^{(t)}= 4\varepsilon_3+(\mu^{''}_{1n}\|\bb_1^{(t)}-\bb_1^*\|_2+\mu^{''}_{2n}\|\bb_2^{(t)}-\bb_2^*\|_2)/\Phi(\cS_3),
    \end{split}
   \end{equation*}
   where $\{\mu_{2n},\mu_{3n}\}, \{\mu_{1n}', \mu_{3n}'\}, \{\mu_{1n}^{''}, \mu_{2n}^{''}\}$ are Lipschitz-gradient parameter which are defined in Condition \ref{con:gradient_stability} and Conditions \ref{con:gradient_stability_b2}-\ref{con:gradient_stability_b3}.
\end{condition}
\begin{remark}
Condition \ref{con:regularization} considers an iterative sequence of regularization parameters. Given reasonable initializations for $\bb_1^{(t)}, \bb_2^{(t)}, \bb_3^{(t)}$, their estimation errors gradually decay when the iteration $t$ increases, which implies that $\lambda_{kn}^{(t)}$ is a decreasing sequence. After sufficiently many iterations, the rate of the $\lambda_{kn}^{(t)}$ will be bounded by the statistical error $\varepsilon_k$, for $k=1,2,3$. This agrees with the theory of high-dimensional regularized M-estimator in that suitable tuning parameter should be proportional to the target estimation error \citep{Wain2014}. Such iterative turning procedure plays a critical role in controlling statistical and optimization error, and has been commonly used in other high-dimensional non-convex optimization problems \citep{wang2014optimal, yi2015regularized}. 
\end{remark}
Finally, we present a general condition on the penalty term.
\begin{condition}[(Decomposable Penalty)]\label{con:decomposable}
  Given a space $\cS$, a norm-based penalty $\cP$ is assumed to be decomposable with respect to $\cS$ such that it satisfies $\cP(\bu+\bv) = \cP(\bu)+\cP(\bv)$ for any $\bu\in \cS$ and $\bv\in \cS^{\perp}$, where $\cS^{\perp}$ is the complement pf $\cS$.
\end{condition}
As shown in \cite{negahban2011}, a broad class of penalties satisfies the decomposable property, such as lasso, ridge, fused lasso, group lasso penalties. Next theorem quantifies the error of one-step update for the estimator coming Algorithm \ref{alg:main}.
\begin{theorem}[Contraction Property]\label{thm:para}
 Suppose Conditions \ref{con:gradient_stability},\ref{con:strongly_convex},\ref{con:stat_error},\ref{con:regularization}, \ref{con:gradient_stability_b2}-\ref{con:stat_error_b3} hold. 
 Assume the update at $t$-th iteration of Algorithm \ref{alg:main}, $\bb_1^{(t)}, \bb_2^{(t)}, \bb_3^{(t)}$ fall into sparse balls $\cB_{\alpha_1, s_1}(\bb_1^*), \cB_{\alpha_2, s_2}(\bb_2^*), \cB_{\alpha_3, s_3}(\bb_3^*)$ respectively, where $\alpha_1,\alpha_2,\alpha_3$ are some constants. Define $\cE^{(t)} = \|\bb_1^{(t)}-\bb_1^*\|_2^2+\|\bb_2^{(t)}-\bb_2^*\|_2^2+\|\bb_3^{(t)}-\bb_3^*\|_2^2$. There exists absolute constant $C_0>1$ such that, the estimation error of the update at the $t+1$-th iteration satisfies,
\begin{eqnarray}\label{eqn:one_step_error}
  \cE^{(t+1)}\leq \rho\cE^{(t)}+ C_0\Big(\frac{\varepsilon_1^2\Phi(\cS_1)^2}{\gamma_{1n}^2}+\frac{\varepsilon_2^2\Phi(\cS_2)^2}{\gamma_{2n}^2}+\frac{ \varepsilon_3^2\Phi(\cS_3)^2}{\gamma_{3n}^2}\Big),
\end{eqnarray}
  with probability at least $1-3(\delta_1+\delta_2+\delta_3)$. Here, $\rho$ is the contraction parameter defined as
  \begin{equation*}
\rho =C_1\max\{\mu_{1n}^{'2}, \mu_{1n}^{''2},\mu_{2n}^2, \mu_{2n}^{''2},\mu_{3n}^2, \mu_{3n}^{'2}\}/\min\{\gamma_{1n}^2, \gamma_{2n}^2, \gamma_{3n}^2\},
\end{equation*}
where $C_1$ is some constant.
\end{theorem}

Theorem~\ref{thm:para} demonstrates the mechanism of how the estimation error improves in the one-step update. When the the contraction parameter $\rho$ is strictly less than 1 (we will prove that it holds for certain class of basis functions and penalties in next section), the first term of RHS in \eqref{eqn:one_step_error} will gradually go towards zero and the second term will be stable. The contraction parameter $\rho$ is roughly the ratio of Lipschitz-gradient parameter and the strongly convex parameter. Similar formulas of contraction parameter frequently appears in the literature of statistical guarantees for low/high-dimensional non-convex optimization \citep{BWY15, hao2017simultaneous}.
\begin{remark}
\cite{yi2015regularized} provided similar optimization and statistical guarantee for regularized EM algorithm based on mixture model. However, the source of non-convexity in the mixture model comes from the  latent variable while ours comes from the bi-convex structure in the low-rank model. Thus their analysis is not directly applicable to our case due to different verification of sparse-strong-convexity condition.
\end{remark}
\subsection{Application to STAR estimator}
\label{sec:thm_STAR}

In this section, we apply the general contract property in Theorem \ref{thm:para} to STAR estimator with B-spline basis functions. The formal definition of B-spline basis function is defined in Section \ref{subsec:B-spline} of the appendix. To ensure Conditions \ref{con:gradient_stability}-\ref{con:strongly_convex} and \ref{con:stat_error} are satisfied, in our STAR estimator we require conditions on the nonparametric component, the distribution of tensor covariate and the noise distribution.
 
\begin{condition}[(Function Class)]\label{con:nonpara_component}
   Each nonparametric component in \eqref{eqn:STAR} is assumed to belong to the function class $\cH$ defined as follows,
  \begin{equation}\label{eqn:function_class}
  \cH = \Big\{g(\cdot):|g^{(r)}(s) - g^{(r)}(t)|\leq C |s-t|^{\alpha}, \ \text{for} \ s,t\in[a,b]\Big\},
  \end{equation}
  where $r$ is the order of the derivative.
  Let $\kappa = r+\alpha>0.5$ be the smoothness parameter of function class $\cH$. 
    For $j\in[p_1], k\in[p_2], l\in[p_3]$, there is a constant $c_f>0$ such that $\min_{jkl}\|f_{jkl}^*(x)\|_2\geq c_f$ and $\mathbb E(f_{jkl}^*([\cX]_{jkl})) =0$. Each component of the covariate tensor $\cX$ has an absolutely continuous distribution and its density function is bounded away from zero and infinitely on $C$.
\end{condition}
 Condition \ref{con:nonpara_component} is classical for nonparametric additive model \citep{stone1985additive,huang2010variable, fan2011nonparametric}. Such condition is required to optimally estimate each individual additive component in $\ell_2$-norm. 

\begin{condition}[(Sub-Gaussian Noise)]\label{con:noise}
  The noise $\{\epsilon_i\}_{i=1}^n$ are i.i.d. randomly generated with mean 0 and bounded variance $\sigma^2$. Moreover, $(\epsilon_i/\sigma)$ is sub-Gaussian distributed, i.e., there exists constant $C_{\epsilon}>0$ such that $\|(\epsilon_i/\sigma)\|_{\phi_2} := \sup_{p\geq 1} p^{-1/2}(\mathbb{E}|\epsilon_i/\sigma|^p)^{1/p}\leq C_{\epsilon}$, and independent of tensor covariates $\{\cX_i\}_{i=1}^n$.
\end{condition}

\begin{condition}[(Parameter Space)]\label{con:signal}
  We assume the absolute value of maximum entry of $(\bb_1^{*\top}, \bb_2^{*\top},\bb_2^{*\top})$ is upper bounded by some positive constant $c^*$, and the absolute value of minimum non-zero entry of $(\bb_1^{*\top}, \bb_2^{*\top},\bb_2^{*\top})$ is lower bounded by some positive constant  $c_*$. Here, $c^*,c_*$ not depending on $n,p$. Moreover, we assume the CP-rank $R$ and sparsity parameters $s_1, s_2, s_3$ are bounded by some constants. 
\end{condition}
\begin{remark}
The condition of bounded elements of tensor coefficient widely appears in tensor literature \citep{anandkumar2014tensor, sun2017provable}. Here the bounded tensor rank condition is imposed purely for simplifying the proofs and this condition is possible to relax to allow slowly increased tensor rank \citep{sun2018dynamic}. The fixed sparsity assumption is also required in the vector nonparametric additive regression \citep{huang2010variable}. To relax it, \cite{meier2009high} considered a diverging sparsity scenario but required a compatibility condition which was hard to verify in practice. Thus, in this paper we consider a fixed sparsity case and leave the extension of diverging sparsity as future work. 
\end{remark}
Since the penalized empirical risk minimization \eqref{def:ERM} is a highly non-convex optimization, we require some conditions on the initial update in Algorithm \ref{alg:main}.

\begin{condition}[(Initialization)]\label{con:initial}
The initialization of $\bb_1, \bb_2, \bb_3$ is assumed to fall into a sparse constant ball centered at $\bb_1^*, \bb_2^*, \bb_3^*$, saying $\bb_1^{(0)}\in\cB_{\alpha_1,s_1}(\bb_1^*), \bb_2^{(0)}\in\cB_{\alpha_2,s_1}(\bb_2^*), \bb_3^{(0)}\in\cB_{\alpha_3,s_1}(\bb_3^*)$, where $\alpha_1,\alpha_2,\alpha_3$ are some constants that are not diverging with $n$ or $p$.
\end{condition}

\begin{remark}
Similar initialization conditions have been widely used in tensor decomposition \citep{sun2017provable, sun2018dynamic}, tensor regression \citep{suzuki2016minimax, sun2017store}, and other non-convex optimization \citep{wang2014optimal, yi2015regularized}. Once the initial values fall into the sparse ball, the contract property and group lasso ensure that the successive updates also fall into a sparse ball. Another line of work considers to design spectral methods to initialize certain simple non-convex optimization, such as matrix completion \citep{ma2017implicit} and tensor sketching \citep{hao2018sparse}. The success of spectral methods heavily relies on a simple non-convex geometry and explicit form of high-order moment calculation, which is easy to achieve in previous work \citep{ma2017implicit,hao2018sparse} by assuming a Gaussian error assumption. However, the design of spectral method in our tensor additive regression is substantially harder since the high-order moment calculation has no explicit form in our context. We leave the design of spectral-based initialization as future work.
\end{remark}

Finally, we state the main theory on the estimation error of our STAR estimator with B-spline basis functions and a group lasso penalty.
\begin{theorem}\label{thm:star}
  Suppose Conditions \ref{con:regularization}, \ref{con:nonpara_component}-\ref{con:signal}, \ref{con:initial} hold and consider the class of normalized B-spline basis functions defined in \eqref{def:B_spline} and group-lasso penalty defined in \eqref{eqn:group_lasso}. If one chooses the number of spline series $d_n\asymp n^{\tfrac{1}{2\kappa+1}}$ and the tuning parameter $\lambda_{1n}^{(t)},\lambda_{2n}^{(t)}, \lambda_{3n}^{(t)}$ as defined in Condition \ref{con:regularization} with generic parameters specified in Lemmas \ref{lemma:oe}-\ref{lemma:sc}, with probability at least $1-C_0(t+1)(Rsn^{-\tfrac{2\kappa}{2\kappa+1}}+1/p)$, we have 
  \begin{equation}\label{eqn:final_estimation_error}
  \cE^{(t+1)}\leq\underbrace{\rho^{t+1}\cE^{(0)}}_{\text{optimization error}}+ \underbrace{\frac{C_1R^2}{1-\rho}n^{-\tfrac{2\kappa-1}{2\kappa+1}}\log (pd_n)}_{\text{statistical error}},
  \end{equation}
  where $0<\rho\leq 1/2$ is a contraction parameter, and $\kappa$ is the smoothness parameter of function class $\cH$ in \eqref{eqn:function_class}. Note that $C_1$ may involve a smaller order term that is negligible asymptotically. Consequently, when the total number of iterations is no smaller than 
  $$
  T^*=\log\left(\frac{1-\rho}{C_1\cE^{(0)}}\frac{n^{\frac{2\kappa-1}{2\kappa+1}}}{\log (pd_n)}\right) / \log(1/\rho),
  $$ and the sample size $n\geq C_2(\log p)^{\tfrac{2\kappa +1}{2\kappa -1}}$ for sufficiently large $C_2$, we have
  \begin{equation*}
  \cE^{(T^*)}\leq \frac{2C_1R^2}{1-\rho}n^{-\frac{2\kappa-1}{2\kappa+1}}\log (pd_n),
  \end{equation*}
  with probability at least $1-C_0(T^*+1)(Rsn^{-\tfrac{2\kappa}{2\kappa+1}}+1/p)$.
\end{theorem}

The non-asymptotic estimation error bound \eqref{eqn:final_estimation_error} consists of two parts: an optimization error which is incurred by the non-convexity and a statistical error which is incurred by the observation noise and the spline approximation. Here, optimization error decays geometrically with the iteration number $t$, while the statistical error remains the same when $t$ grows. When the tensor covariate is of order-one, i.e., a vector covariate, the overall optimization problem reduces to classical vector nonparametric additive model. In that case, we do not have the optimization error $(\rho^{t+1}\cE^{(0)})$ any more since the whole optimization is convex, and the statistical error term matches the state-of-the-art rate in \cite{huang2010variable}. 

Lastly, let us define $\hat{\cT}(\cX) = \sum_{h=1}^{d_n}\sum_{r=1}^R\langle \bbeta_{1hr}^{(T^*)}\circ\bbeta_{2hr}^{(T^*)}\circ\bbeta_{3hr}^{(T^*)}, \cF_h(\cX) \rangle$ as a final estimator of the target function $\cT^*(\cX)$.  For any function $f$ on $[a, b]$, we define its $\ell_2(P)$ norm by $\|f\|_2 = \sqrt{\int_{a}^bf^2(x)dP(x)}$. The following corollary provides the final error rate for the estimation of tensor additive nonparametric function $\cT^*(\cX)$. It incorporates the approximation error of nonparametric component incurred by the B-spline series expansion, and the estimation error of unknown tensor parameter incurred by noises.
\begin{corollary}\label{cor:func}
    Suppose Conditions \ref{con:regularization}, \ref{con:nonpara_component}-\ref{con:signal}, \ref{con:initial} hold and the number of iterations $t$ as well as the sample size $n$ satisfy the requirement in Theorem \ref{thm:star}. Assume the non-zero elements in $\cT^*(\cdot)$ are the same as $\sum_{h=1}^{d_h}\langle \cB_h^*, \cF_h(\cdot)\rangle$. Then, the final estimator satisfies
  \begin{equation*}
  \big\|\hat{\cT}-\cT^*\big\|_2^2= \cO_p\Big(n^{-\tfrac{2\kappa}{2\kappa+1}}\log (pd_n)\Big).
  \end{equation*}
\end{corollary}

\section{Simulations}\label{sec:simu}
In this section, we carry out intensive simulation studies to evaluate the performance of our STAR method, and compare it with existing competing solutions including the tensor linear regression (TLR) \citep{zhou2013}, the Gaussian process based nonparametric method (GP) \citep{kanagawa2016gaussian}, and the nonlinear tensor regression via alternative minimization procedure (AMP) \citep{suzuki2016minimax}. We find that STAR enjoys better performance both in terms of prediction accuracy and computational efficiency. 

Throughout our numerical studies, the natural cubic splines with B-spline basis are used in STAR with the degree fixed to be five, which amounts to having four inner knots. For both STAR and TLR, five-fold cross-validation is employed to select the best pair of the tuning parameters $R$ and $\lambda$, where the tensor rank $R$ is chosen from $\{2, 3\}$ and $\lambda$ is selected from a sequence that is uniformly distributed on the logarithm scale in an interval $[10^{-5}, 1]$. For GP and AMP, as suggested by \cite{kanagawa2016gaussian}, the Gaussian kernel is used and the bandwidth is set to be 100; five-fold cross-validation is used to select $\lambda$, where $\lambda$ is selected from the same range that is used for TLR and STAR.

\subsection{Low-rank covariate structure}
\label{sec:low_rank}
The simulated data are generated based on the following model,
\begin{equation*}
y_i = \cT^*(\cX_i) + \sigma \epsilon_i, \ i = 1, \ldots, n,
\end{equation*}
where $\epsilon_i \sim \mathrm{N}(0,1)$. For each observation, $y_i\in\mathbb{R}$ is the response and $\cX_i \in \mathbb{R}^{p_1 \times p_2}$ is the two-way tensor (matrix) covariate. We fix $p_2=8$, and we vary $n$ from $\{400, 600\}$, $p_1$ from $\{20, 50, 100\}$, and $\sigma$ from $\{0.1, 1\}$. We assume that there are $10$ and $4$ important features along the first and second way of $\cX$, respectively.

Since both GP and AMP models require a low-rank structure on the tensor covariates, in this simulation we consider the low-rank covariate case which favors their models. For each $i = 1, \ldots, n$, we consider $\cX_i = \bx^{(1)}_i \circ \bx^{(2)}_i$, where the elements of $\bx^{(1)}_i \in \mathbb{R}^{p_1}$ and $\bx^{(2)}_i \in \mathbb{R}^{p_2}$ are independently and identically distributed from uniform distribution. Following the additive model that is considered in \cite{Ravikumar2009}, we generate samples according to
$$
y_i = \sum_{j=1}^{10} \sum_{k=1}^4 \cT^*_j \left([\bx ^{(1)}_i]_j \right) \cdot \cT^*_k \left([\bx ^{(2)}_i]_k\right) + \sigma \epsilon_i, \ i=1, \ldots, n,
$$
where the nonlinear functions $\cT^*_j$ and $\cT^*_k$ are given by 
$$
\cT^*_j(x) = 
\begin{cases}
-\sin(1.5x), & \text{if $j$ is odd}, \\
x^3 + 1.5(x - 0.5)^2, & \text{if $j$ is even}, \\
\end{cases}
\textrm{~and~}\cT^*_k(x) = 
\begin{cases}
-\phi(x, 0.5, 0.8^2), & \text{if $k$ is odd}, \\
\sin\{ \exp(-0.5x)\}, & \text{if $k$ is even}.\\
\end{cases}
$$
Here $\phi(\cdot, 0.5, 0.8^2)$ is the probability density function of the normal distribution $\mathrm{N}(0.5, 0.8^2)$.

Table~\ref{tab:low_rank} compares the mean squared error (MSE) of all four models, where MSE is assessed on an independently generated test data of $2,000$ samples. The STAR model shows the lowest MSE in all cases. As expected, the TLR model has unsatisfactory performance, as the true regression model is non-linear. The two nonparametric tensor regression models GP and AMP can capture partial nonlinear structures, however, our method still outperform theirs.

\begin{table}[!t]
\caption{MSE of simulated data with low-rank covariate structure.}
\label{tab:low_rank}
\centering
\vspace{0.3cm}
\resizebox{0.6\textwidth}{!}{%
\ra{1.1}
\begin{threeparttable}
\begin{tabular}{rrrrrrrrrrrrrrrrrr}
\toprule
$(n,\sigma)$&model&\multicolumn{2}{c}{$p_1=20$} \phantom{} & \multicolumn{2}{c}{$p_1=50$}\phantom{}&
 \multicolumn{2}{c}{$p_1=100$}\\
  \midrule
$(400, 0.1)$ & \textbf{STAR} & 0.51 & (0.01) & 0.53 & (0.01) & 0.50 & (0.01) \\ & TLR &2.03 & (0.17)  & 2.63 & (0.17)  & 3.16 & (0.48) \\ & AMP &1.02 & (0.02)  & 1.01 & (0.02)  & 1.01 & (0.02) \\ & GP &1.02 & (0.01)  & 1.03 & (0.03)   & 1.02 & (0.01) \\
$(600, 0.1)$ & \textbf{STAR} & 0.50 & (0.01) & 0.50 & (0.01) & 0.52 & (0.01) \\ & TLR &2.11 & (0.09)  & 2.81 & (0.14)  & 3.19 & (0.21) \\ & AMP &0.99 & (0.02)  & 1.01 & (0.00)  & 1.00 & (0.02) \\ & GP &0.99 & (0.02) & 1.01 & (0.01) & 1.02 & (0.01) \\
$(400, 1)$  & \textbf{STAR} & 1.54 & (0.02) & 1.59 & (0.03) & 1.56 & (0.01) \\ & TLR &2.29 & (0.17)  & 3.18 & (0.40)  & 4.91 & (0.56) \\ & AMP &1.98 & (0.02)  & 2.01 & (0.01)  & 2.06 & (0.03) \\ & GP &2.05 & (0.04) & 2.04 & (0.03) & 2.07 & (0.03) \\
$(600, 1)$ & \textbf{STAR} & 1.55 & (0.01) & 1.53 & (0.01) & 1.54 & (0.01) \\ & TLR &2.89 & (0.39)  & 3.90 & (0.36)  & 4.69 & (0.46) \\ & AMP &2.02 & (0.02)  & 2.03 & (0.03)  & 2.04 & (0.03) \\ & GP &2.02 & (0.01) & 2.04 & (0.04) & 2.06 & (0.03) \\
   \bottomrule
\end{tabular}
\begin{tablenotes}\footnotesize
\item[*] The simulation compares sparse tensor additive regression (STAR), tensor linear regression (TLR), alternative minimizing procedure (AMP), and Gaussian process (GP). The reported errors are the medians over 20 independent runs with the standard error given in parentheses. 
\end{tablenotes}
\end{threeparttable}
}
\end{table}

Next, we investigate the computational costs of all four methods. Table~\ref{tab:time} compares the computation time in the example with $n = 400$ and $\sigma = 0.1$. The results of other scenarios are similar and hence omitted. All the computation time includes the model fitting and the model tuning using five-fold cross-validation. Overall our STAR method is as fast as AMP and is much faster than GP. When the tensor dimension is small, $p_1 = 20$, the linear model TLR is the fastest one, however, its computation cost dramatically increase when the dimension $p_1$ increases, and is even slower than other nonparametric models when $p_1 = 100$. On the other hand, the computation time of our model is less sensitive to the dimensionality and is even faster than TLR when $p_1$ is large. This indicates the importance of fully exploiting the low-rankness and sparsity structures in order to improve computational efficiency. 



\begin{table}[!t]
\caption{The computation time of all methods in Section~\ref{sec:low_rank} with $n=400$ and $\sigma= 0.1$.}
\label{tab:time}
\centering
\vspace{0.3cm}
\resizebox{0.4\textwidth}{!}{%
\ra{1.1}
\begin{threeparttable}
\begin{tabular}{rrrrrrrrrrrrrrrrrr}
\toprule
&$p_1$&\multicolumn{1}{c}{$20$} \phantom{} & \multicolumn{1}{c}{$50$}\phantom{}&
 \multicolumn{1}{c}{$100$}\\
  \midrule
& \textbf{STAR} & 353.30 & 227.65 & 391.90\\ & TLR & 156.19 & 738.96 & 1803.49\\ & AMP &  330.74 & 336.10 & 341.15 \\ & GP & 1841.51 & 1913.27 & 1792.50\\
   \bottomrule
\end{tabular}
\begin{tablenotes}\footnotesize
\item[*] All the time include five-fold cross-validation procedures to tune the parameters. The results are averaged over 20 independent runs. The experiment was conducted using a single processor Inter(R) Xeon(R) CPU E5-2600@2.60GHz.
\end{tablenotes}
\end{threeparttable}
}
\end{table}

\subsection{General covariate structure}\label{sec:full_rank}

The settings are similar as that in Section \ref{sec:low_rank}, except that the covariate is not low-rank. Here, the elements of $\cX_i \in \mathbb{R}^{p_1 \times p_2}$ are independently and identically distributed from uniform distribution. In particular, we generate $n$ observations according to 
$$
y_i = \sum_{j=1}^{10} \sum_{k=1}^{4} \cT^*_{jk}([\cX_i]_{jk}) + \sigma \epsilon_i,  \ i = 1, \ldots, n,
$$
where
\begin{equation*}
\cT^*_{jk}(x) =
\begin{cases}
-\sin(1.5x), & \text{ if $j$ is odd and $k$ is odd}, \\
x^3 + 1.5(x - 0.5)^2, & \text{ if $j$ is even and $k$ is odd}, \\
-\phi(x, 0.5, 0.8^2), & \text{ if $j$ is odd and $k$ is even}, \\
\sin\{ \exp(-0.5x)\}, & \text{ if $j$ is even and $k$ is even}.\\
\end{cases}
\end{equation*}
To run GP and AMP models, we perform singular value decomposition in order to meet the requirements of the low-rank tensor inputs. 

The comparisons of the MSE are summarized in Table~\ref{tab:full_rank}: the MSE of the STAR model is much lower than that of all the other three models. Similar to the experiment in Section \ref{sec:low_rank}, the large MSE of TLR is attributed to the incapability of capturing the nonlinear relationship in the additive model. In this example, the GP and AMP models deliver relatively large MSE because the assumption of low-rank covariate structure is violated. Importantly, the MSE of our STAR model decreases, as the sample size increases or the noise level $\sigma$ decreases. These observations align with our theoretical finding in Theorem \ref{thm:star}.

To test the adaptability of STAR, we further consider a linear data-generating model:
$$
y_i = \sum_{j=1}^{10} \sum_{k=1}^{4} \cT^*_{jk}([\cX_i]_{jk}) + \sigma \epsilon_i,  \ i = 1, \ldots, n,
$$
where
\begin{equation*}
\cT^*_{jk}(x) =
\begin{cases}
0.5 x, & \text{ if $j$ is odd}, \\
x, & \text{ if $j$ is even}. \\
\end{cases}
\end{equation*}

We use $n = 400$ and $\sigma=1$ for illustration and summarize the result in Table~\ref{tab:linr}. We observe TLR delivers the lowest MSE and STAR is the second best when $p_1 = 20$ and $50$. The MSE of STAR is the lowest when $p_1 = 100$. Thus the STAR model is competitive with TLR in general when the data generating model is truly linear.

\begin{table}[!t]
\caption{MSE of simulated data with general covariate structure.}
\label{tab:full_rank}
\centering
\vspace{0.3cm}
\resizebox{0.6\textwidth}{!}{%
\ra{1.1}
\begin{threeparttable}
\begin{tabular}{rrrrrrrrrrrrrrrrrr}
\toprule
$(n,\sigma)$&model&\multicolumn{2}{c}{$p_1=20$} \phantom{} & \multicolumn{2}{c}{$p_1=50$}\phantom{}&
 \multicolumn{2}{c}{$p_1=100$}\\
  \midrule
$(400, 0.1)$ & \textbf{STAR} & 2.30 & (0.06) & 3.34 & (0.22) & 5.04 & (0.33) \\ & TLR &40.62 & (0.53)  & 53.94 & (0.71)  & 92.31 & (1.60) \\ & AMP &41.09 & (0.31)  & 40.78 & (0.35)  & 40.97 & (0.30) \\ & GP &41.62 & (0.24) & 41.35 & (0.38) & 41.18 & (0.22) \\
$(600, 0.1)$ & \textbf{STAR} & 1.76 & (0.04) & 2.14 & (0.11) & 2.53 & (0.12) \\ & TLR &37.12 & (0.63)  & 43.74 & (0.85)  & 58.75 & (0.88) \\ & AMP &40.39 & (0.30)  & 40.35 & (0.35)  & 41.00 & (0.37) \\ & GP &40.65 & (0.66) & 40.57 & (0.64) & 41.28 & (0.44) \\
$(400, 1)$ & \textbf{STAR} & 3.98 & (0.11) & 5.28 & (0.26) & 7.17 & (0.50) \\ & TLR &42.12 & (0.51)  & 56.47 & (0.78)  & 94.09 & (2.38) \\ & AMP &42.00 & (0.38)  & 42.40 & (0.47)  & 41.51 & (0.28) \\ & GP &42.56 & (0.36) & 42.19 & (0.37) & 42.08 & (0.30) \\
$(600, 1)$ & \textbf{STAR} & 3.13 & (0.05) & 3.63 & (0.13) & 4.10 & (0.11) \\ & TLR &38.06 & (0.59)  & 45.43 & (0.65)  & 60.20 & (1.03) \\ & AMP &41.44 & (0.54)  & 41.99 & (0.45)  & 41.76 & (0.40) \\ & GP &41.31 & (0.50) & 41.87 & (0.44) & 42.49 & (0.53) \\
   \bottomrule
\end{tabular}
\begin{tablenotes}\footnotesize
\item[*] The simulation compares sparse tensor additive regression (STAR), tensor linear regression (TLR), alternative minimizing procedure (AMP), and Gaussian process (GP). The reported errors are the medians over 20 independent runs with the standard error given in parentheses.  
\end{tablenotes}
\end{threeparttable}
}
\end{table}

\begin{table}[!t]
\caption{MSE of simulated data from linear models.}
\label{tab:linr}
\centering
\vspace{0.3cm}
\resizebox{0.5\textwidth}{!}{%
\ra{1.1}
\begin{threeparttable}
\begin{tabular}{rrrrrrrrrrrrrrrrrr}
\toprule
&model&\multicolumn{2}{c}{$p_1=20$} \phantom{} & \multicolumn{2}{c}{$p_1=50$}\phantom{}&
 \multicolumn{2}{c}{$p_1=100$}\\
  \midrule
   & \textbf{STAR} & 1.40 & (0.03) & 1.62 & (0.13) & 1.85 & (0.06) \\ & TLR & 1.19 & (0.01)  & 1.52 & (0.02)  & 2.42 & (0.04) \\ & AMP & 2.07 & (0.02)  & 2.08 & (0.02)  & 2.09 & (0.03) \\ & GP & 2.10 & (0.01) & 2.10 & (0.01) & 2.10 & (0.01) \\
   \bottomrule
\end{tabular}
\begin{tablenotes}\footnotesize
\item[*] The simulation compares sparse tensor additive regression (STAR), tensor linear regression (TLR), alternative minimizing procedure (AMP), and Gaussian process (GP). The reported errors are the medians over 20 independent runs with the standard error given in parentheses. 
\end{tablenotes}
\end{threeparttable}
}
\end{table}

\subsection{Three-way covariate structure}
We next extend the previous simulations to a three-way covariate structure. We consider two cases in this section. In the first case, we generate the tensor covariate $\mathcal{X}_i \in \mathbb{R}^{p_1 \times p_2 \times 2}$ whose elements are from i.i.d. uniform distribution, and then generate the response $y_i \in \mathbb{R}$ from
$$
\text{case 1: }y_i = \sum_{j=1}^{10} \sum_{k=1}^{4} \left[\sin(\cT^*_{jk1}([\cX_i]_{jk1})) + \log|\cT^*_{jk2}([\cX_i]_{jk2})|\right] + \sigma \epsilon_i,  \ i = 1, \ldots, n,
$$
where 
$\cT^*_{jkl}$ with $l=1,2$ is defined the same as the expression \eqref{eqn:simu_setting}. In the second case, we generate the response from
\[
\text{case 2: }y_i = \sin \left(\sum_{j=1}^{10} \sum_{k=1}^{4} \cT^*_{jk}([\cX_i]_{jkl}) \right) + \log\left|\sum_{j=1}^{10} \sum_{k=1}^{4} \cT^*_{jk}([\cX_i]_{jkl})\right| + \sigma \epsilon_i,  \ i = 1, \ldots, n.
\]
It is worth noting that case 1 uses an additive model while case 2 does not. Therefore, the additive model assumption in our STAR method is actually mis-specified in case 2.

\begin{table}[!t]
\caption{MSE of simulated data with three-way tensor covariates.}
\label{tab:three_mode2}
\centering
\resizebox{0.6\textwidth}{!}{%
\ra{1.1}
\begin{threeparttable}
\begin{tabular}{rrrrrrrrrrrrrrrrrrrrrrr}
\toprule
&& \multicolumn{2}{c}{$p_1=20$} \phantom{} & \multicolumn{2}{c}{$p_1=50$}\phantom{}&
 \multicolumn{2}{c}{$p_1=200$}\\
  \midrule
 \multicolumn{8}{c}{\textbf{Case 1}}\\
 $(600, 0.1)$& \textbf{STAR} & 0.34 & (0.03) & 1.33 & (0.23) & 1.10 & (0.03)\\ & TLR &10.69 & (0.06) & 10.86 & (0.14) & 13.47 & (0.25)\\
$(600, 1)$ & \textbf{STAR} & 1.45 & (0.02) & 1.71 & (0.20) & 1.99 & (0.01)\\ & TLR &11.93 & (0.09) & 12.45 & (0.14) & 18.14 & (0.95)\\
$(1000, 0.1)$ & \textbf{STAR} & 0.37 & (0.02) & 0.43 & (0.03) & 1.24 & (0.25)\\ & TLR &10.82 & (0.10) & 10.77 & (0.12) & 11.68 & (0.11)\\
$(1000, 1)$ & \textbf{STAR} & 1.40 & (0.03) & 1.47 & (0.02) & 2.02 & (0.03)\\ & TLR &11.90 & (0.06) & 11.98 & (0.15) & 13.58 & (0.19)\\
 \multicolumn{8}{c}{\textbf{Case 2}}\\
 $(600, 0.1)$& \textbf{STAR} & 0.78 & (0.01) & 2.07 & (1.10) & 1.01 & (0.00)\\ & TLR &13.11 & (0.11) & 13.11 & (0.17) & 16.62 & (0.33)\\
 $(600, 1)$& \textbf{STAR} & 2.07 & (0.08) & 2.01 & (0.02) & 1.99 & (0.02)\\ & TLR &14.45 & (0.17) & 14.33 & (0.14) & 20.73 & (0.66)\\
$(1000, 0.1)$ & \textbf{STAR} & 0.75 & (0.00) & 0.78 & (0.05) & 1.35 & (0.17)\\ & TLR &13.15 & (0.22) & 12.98 & (0.17) & 13.52 & (0.23)\\
$(1000, 1)$ & \textbf{STAR} & 1.76 & (0.02) & 2.13 & (0.12) & 1.99 & (0.02)\\ & TLR &14.15 & (0.29) & 14.05 & (0.19) & 15.57 & (0.27)\\
   \bottomrule
\end{tabular}
\begin{tablenotes}\footnotesize
\item[*] The simulation compares sparse tensor additive regression (STAR) and tensor linear regression (TLR). The reported errors are the medians over 20 independent runs, and the standard error of the medians are given in parentheses. All the methods use five-fold cross-validation procedures to tune the parameters.
\end{tablenotes}
\end{threeparttable}
}
\end{table}

Since the softwares of AMP and GP models for three-way tensor covariates are not available, in this simulation we compare our STAR model only with TLR. We vary the sample size $n \in \{600, 1000\}$, the first-way dimension $p_1 \in \{20, 50, 200\}$, the noise level $\sigma \in \{0.1, 1\}$, and fix the second-way dimension $p_2=10$. Similar to previous simulations, we assume that there are $10$, $4$, and $2$ important features along the three modes of the tensor $\cX_i$, respectively. As shown in Table~\ref{tab:three_mode2}, the MSE of the STAR model is consistently lower than that of TLR, even in the case when the additive model is mis-specified.



\section{An Application to Online Advertising}\label{sec:real}

In this section, we apply the STAR model to click-through rate (CTR) prediction in online advertising. The CTR is defined to be the ratio between the number of clicks and the number of impressions (ad views). In this study, we are interested in predicting the overall CTR, which is the average CTR across different ad campaigns. The overall CTR is an effective measure to evaluate the performance of online advertising. A low overall CTR usually indicates that the ads are not effectively displayed or the wrong audience is being targeted. As a reference, the across-industry overall CTR of display campaigns in the United States from April 2016 to April 2017 is $0.08\%$.\footnote{The data are from \url{http://www.richmediagallery.com/learn/benchmarks}.} Importantly, the CTR is also closely related to the revenue. Define the effective revenue per mile (eRPM) to be the amount of revenue from every 1000 impressions, and we have $\mathrm{eRPM} = 1000 \times \mathrm{CPC} \times \mathrm{CTR}$, where CPC is the cost per click. From this expression, we can see that a good CTR prediction is critical to ad pricing, and the CTR prediction is a highly important task in online advertising. 

We collect 136 ad campaign data during 28 days from a premium Internet media company.\footnote{The reported data and results in this section are deliberately incomplete and subject to anonymization, and thus do not necessarily reflect the real portfolio at any particular time.} The data from each day have been aggregated into six time periods and each of the 136 campaigns involves ads delivered via three devices: phone, tablet, and personal computer (PC). In total, we have $224 = 28\times8$ time periods. There are 153 million of users in total, and we divide all the users into two groups, a younger group and an elder group, which are partitioned by the median age. For each time period, we aggregate the number of impressions of 136 advertising campaigns that are delivered on each of three types of devices for each of the two age groups. Denoting the number of impressions by $\mathcal{X}$, each data point has $\mathcal{X}_i \in \mathbb{R}^{136\times 3 \times 2}$, and $i=1,2,\ldots,224$ represents the time period. In this study, we aim to study the relationship between the overall CTR and the three-way tensor covariate of impressions. 

Figure~\ref{fig:nonlinear} delineates the marginal relationship between the overall CTR and the impression of one advertisement that is delivered on phone, tablet, and PC, respectively. In this example, the overall CTR clearly reveals a non-linear pattern across all devices. Moreover, it is generally believed that not all ads have significant impacts on the overall CTR and hence ad selection is an active research area \citep{choi2010using, xu2016lift}. To fulfill both tasks of capturing the nonlinear relationship and selecting important ads, we apply the proposed STAR model to predict the overall CTR. The logarithm transformations are applied to both the CTR and the number of impressions.  We train and tune each method on the data obtained on the first 24 days, and use the remaining data as the test data to assess the prediction accuracy. The MSE of our STAR model is 0.51, which is much lower than 5.44, the MSE of the TLR. This result shows the effectiveness of capturing the non-linear relationship as well as assuming the low-rankness and group sparsity structures both in increasing the CTR prediction accuracy and the algorithm efficiency. The AMP and GP models are not compared due to the lack of implementation for three-way covariates. 

\begin{figure}[t]
\begin{minipage}[b]{\linewidth}
\centering
\includegraphics[width=1.02\textwidth]{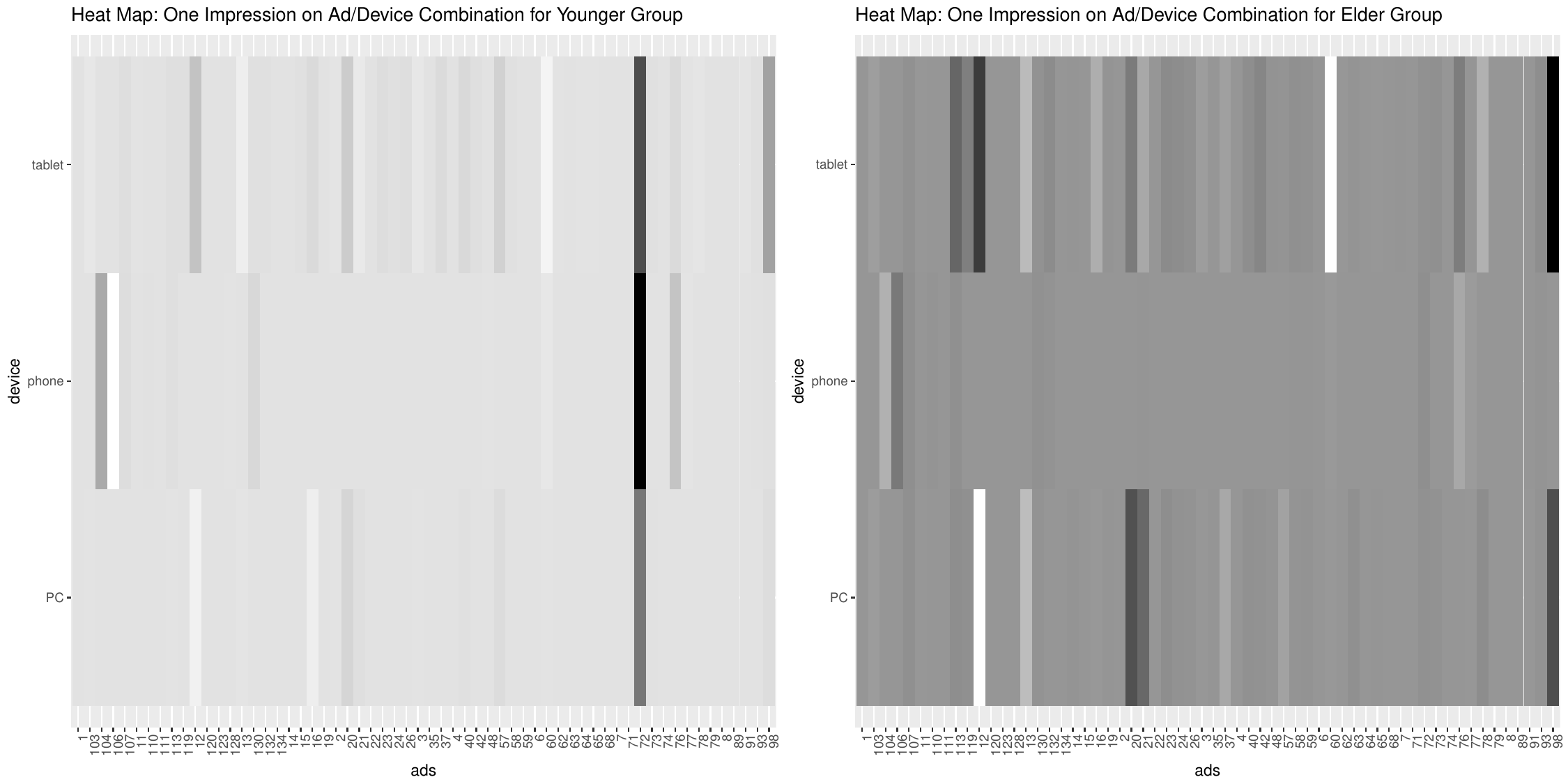}
\caption{\small Heatmaps for the overall CTR. The left panel is the mean change in the overall CTR if the test data have one additional impression on each ad and device combination for the younger group and the right panel is for the elder group. Darker tiles indicate greater positive mean change in the overall CTR and lighter tiles indicate greater negative mean change. The IDs of ads have been renumbered for concerns of confidentiality.}
\label{fig:heatmap1}
\end{minipage}
\end{figure}

\begin{figure}[t]
\begin{minipage}[b]{\linewidth}
\centering
\includegraphics[width=1.02\textwidth]{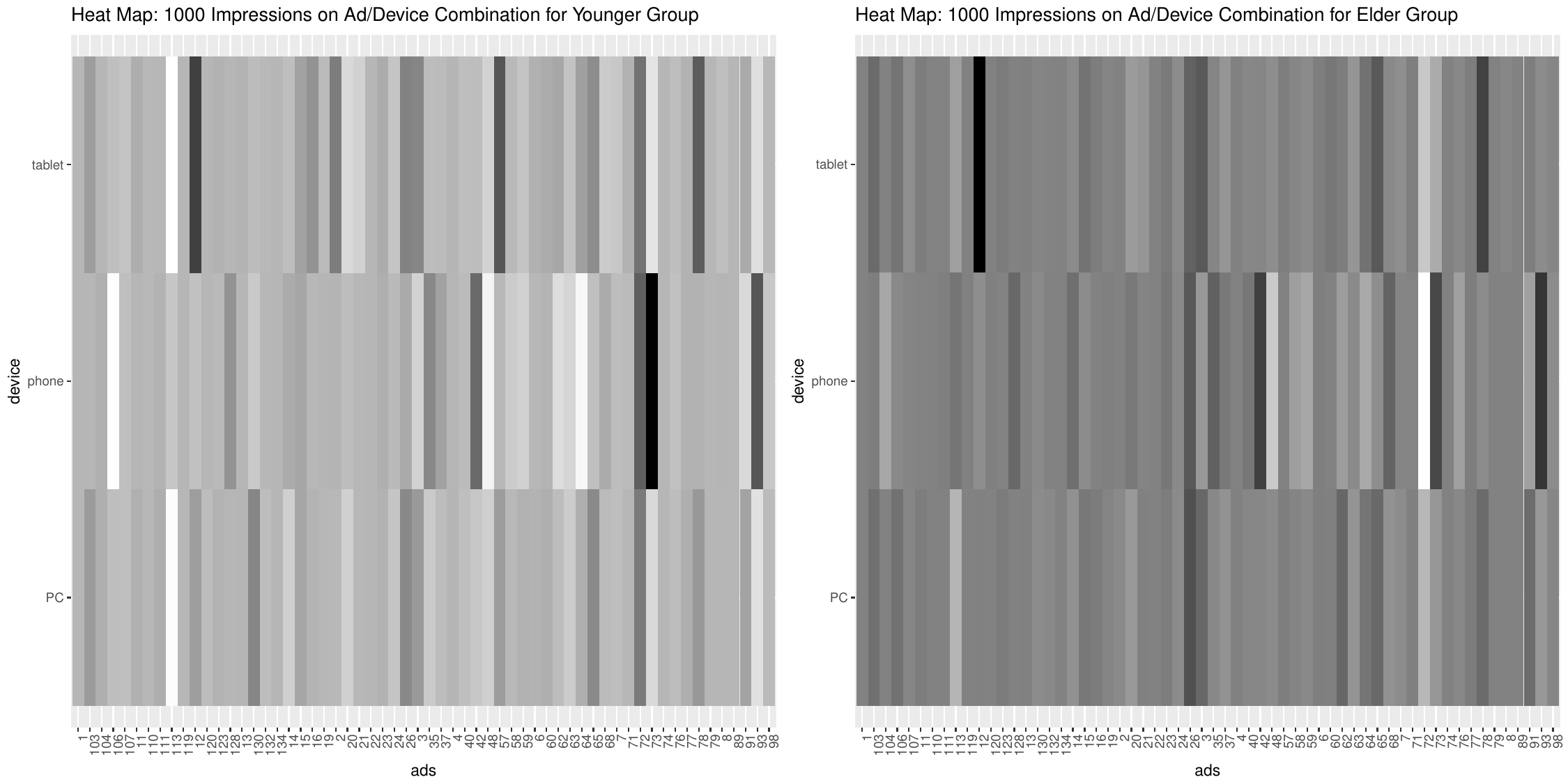}
\caption{\small Heatmaps for the overall CTR. The left panel is the mean change in the overall CTR if the test data have 1000 additional impressions on each ad and device combination for the younger group and the right panel is for the elder group. Darker tiles indicate greater positive mean change in the overall CTR and lighter tiles indicate greater negative mean change. The IDs of ads have been renumbered for concerns of confidentiality.}
\label{fig:heatmap1000}
\end{minipage}
\end{figure}

In terms of ad selection, the STAR model with group lasso penalty selects 60 out of 136 ads, as well as all three devices and two age groups as active variables for the CTR prediction. As a comparison, TLR selects 114 ads, 46 of which are also selected by our STAR method. Besides the prediction on the overall CTR and the ad selection performance, we are also able to see which combination of ad, device, and age group yields the most significant impact on the overall CTR. In the left panel of Figure~\ref{fig:heatmap1}, each tile represents a combination of ad and device for the younger group, and the darkness of the tile implies the sensitivity of the overall CTR associated with one more impression on this combination; the right panel of Figure~\ref{fig:heatmap1} shows the heatmap for the elder group. Displayed on phones of the elder users, the ad with ID 98 has the most positive effect on the overall CTR. Figure~\ref{fig:heatmap1000} is plotted similarly except that the change is due to every 1000 additional impressions on the certain combinations. The overall CTR has the largest growth when 1000 additional impressions are allocated to the ad with ID 73 displaying on phones of the younger users.  The different patterns between Figure~\ref{fig:heatmap1} and \ref{fig:heatmap1000} indicate the nonlinear relationship between the overall CTR and the number of impressions. This result is important for managerial decision making. Under a specific budget, our STAR model facilitates ad placement targeting based on the best ad/device/age combination to maximize the ad revenue.  

\section*{Appendix}
\renewcommand{\thesubsection}{A\arabic{subsection}}
\renewcommand{\theequation}{A\arabic{equation}}
In this appendix, we present the detailed proofs for Theorem \ref{thm:para}, \ref{thm:star} and Corollary \ref{cor:func}.

\subsection{Proof of Theorem \ref{thm:para}}

First, we state a key lemma which quantifies the estimation error of each component individually within one iteration step.  

\begin{lemma}\label{lemma:one-step}
Suppose Conditions \ref{con:gradient_stability}-\ref{con:strongly_convex} and \ref{con:stat_error} hold, and the updates at time $t$ satisfy $\bb_2^{(t)}\in\cB_{\alpha_2, s_2}(\bb_2^*)$, $\bb_3^{(t)}\in\cB_{\alpha_3, s_3}(\bb_3^*)$. Let the penalty $\cP$ fulfills the decomposable property (See Definition \ref{def:dual_norm} for details), and the regularization parameter $\lambda_{1n}^{(t)}\geq 4\varepsilon_1+(\mu_{2n}\|\bb_2^{(t)}-\bb_2^*\|_2+\mu_{3n}\|\bb_3^{(t)}-\bb_3^*\|_2)/\Phi(\cS_1)$ where $\mu_{2n}, \mu_{3n}$ and $\varepsilon_1$ are defined in Condition \ref{con:gradient_stability} and \ref{con:stat_error} respectively. Then the update of $\bb_1$ at time $t+1$ satisfies 
\begin{equation}\label{eqn:generic_update}
    \|\bb_1^{(t+1)}-\bb_1^*\|_2\leq 4\lambda_{1n}^{(t)}\Phi(\cS_1)/\gamma_{1n},
\end{equation}
with probability at least $1-(\delta_1+\delta_2+\delta_3)$, where $\gamma_{1n}$ is defined in Condition \ref{con:strongly_convex} and $\Phi(\cS_1)$ is the support space compatibility constant defined in \eqref{def:support_constant}.
\end{lemma}

\emph{Proof.} For notation simplicity, we will drop the superscript of $\bb_1^{(t)}, \bb_2^{(t)}, \bb_3^{(t)}, \lambda_{1n}^{(t)}$ and replace the superscript of $\bb_1^{(t+1)}, \bb_2^{(t+1)}, \bb_3^{(t+1)}$ by $\bb_1^{+}, \bb_2^{+}, \bb_3^{+}$ in the rest of the proof for Lemma \ref{lemma:one-step}.

First of all,  the loss function \eqref{eqn:rewrite_risk} enjoys a bi-convex structure, in the sense that $\cL(\bb_1, \bb_2, \bb_3)$ is convex in one argument when fixing the other two. Then, given current update $\bb_2, \bb_3$, the penalized alternating minimization with respect to $\bb_1$ takes the form of
$$
\bb_1^{+} = \argmin_{\bb_1}\cL(\bb_1, \bb_2, \bb_3)+\lambda_{1n} \cP(\bb_1).
$$
  As $\bb_1^{+}$ minimizes the loss function, we have 
\begin{eqnarray}\label{eqn:optimality}
\cL(\bb_1^{+}, \bb_2, \bb_3)+\lambda_{1n} \cP(\bb_1^{+})\leq \cL(\bb_1^*,\bb_2, \bb_3)+\lambda_{1n} \cP(\bb_1^*),
\end{eqnarray}
which further implies the following inequality by the convexity of $\cL(\cdot, \bb_2, \bb_3)$,
\begin{equation}\label{ineqn:penalty}
\begin{split}
\lambda_{1n} (\cP(\bb_1^+)-\cP(\bb_1^*))&\leq \cL(\bb_1^*, \bb_2,\bb_3)-\cL(\bb_1^+,\bb_2, \bb_3)\\
&\leq  \underbrace{|\langle\nabla_1\cL(\bb_1^*, \bb_2, \bb_3), \bb_1^+-\bb_1^* \rangle|}_{\text{RHS}}.
\end{split}
\end{equation}
Recall that $\nabla_1\cL$ is the noisy gradient function with respect to $\bb_1$ defined in \eqref{def:gradient}. To separate the statistical error and optimization error, we utilize noiseless gradient function $\nabla_1\tilde{\cL}(\bb_1^*,\bb_2,\bb_3)$ defined in \eqref{def:gradient} as a bridge. The detail decomposition is presented as follows,
\begin{eqnarray*}
  \text{RHS} &\leq& \underbrace{|\langle\nabla_1\cL(\bb_1^*, \bb_2, \bb_3)-\nabla_1\tilde{\cL}(\bb_1^*, \bb_2, \bb_3), \bb_1^+-\bb_1^*\rangle|}_{\text{statistical error}}\\
  &&+\underbrace{|\langle\nabla_1\tilde{\cL}(\bb_1^*, \bb_2, \bb_3)-\nabla_1\tilde{\cL}(\bb_1^*, \bb_2^*, \bb_3^*), \bb_1^+-\bb_1^*\rangle|}_{\text{optimization error}},
\end{eqnarray*}
where $\nabla_1\tilde{\cL}(\bb_1^*, \bb_2^*, \bb_3^*) = 0$. Moreover, based on the decomposability of penalty $\cP$ (See Condition \ref{con:decomposable}),
\begin{eqnarray*}
\text{RHS}  &\leq& \|\nabla_1\cL(\bb_1^*, \bb_2, \bb_3)-\nabla_1\tilde{\cL}(\bb_1^*, \bb_2, \bb_3)\|_{\cP^*} \|\bb_1^+-\bb_1^*\|_{\cP}\\
  &&+\langle\nabla_1\tilde{\cL}(\bb_1^*, \bb_2, \bb_3)-\nabla_1\tilde{\cL}(\bb_1^*, \bb_2, \bb_3^*),\bb_1^+-\bb_1^*\rangle\\
  && + \langle\nabla_1\tilde{\cL}(\bb_1^*, \bb_2, \bb_3^*)-\nabla_1\tilde{\cL}(\bb_1^*, \bb_2^*, \bb_3^*),\bb_1^+-\bb_1^*\rangle,
\end{eqnarray*}
where $\cP^*$ is the dual norm of $\cP$. We write $\cP(\bb_1^+-\bb_1^*)=\|\bb_1^+-\bb_1^*\|_{\cP}$.  In addition, putting \eqref{ineqn:penalty} and Conditions \ref{con:gradient_stability} and \ref{con:stat_error} together, we have
\begin{equation}\label{eqn:L}
\begin{split}
&|\langle\nabla_1\cL(\bb_1^*, \bb_2, \bb_3), \bb_1^+-\bb_1^* \rangle|\\
&\leq \varepsilon_1 \cP(\bb_1^+-\bb_1^*)+ \big(\mu_{2n}\|\bb_2-\bb_2^*\|_2+\mu_{3n}\|\bb_3-\bb_3^*\|_2\big)\|\bb_1^{+}-\bb_1^*\|_2,
\end{split}
\end{equation}
with probability at least $1-(\delta_1+\delta_3)$. Together with \eqref{eqn:optimality},
\begin{eqnarray*}
\lambda_{1n} (\cP(\bb_1^+)-\cP(\bb_1^*))\leq \varepsilon_1 \cP(\bb_1^+-\bb_1^*)+ \big(\mu_{2n}\|\bb_2-\bb_2^*\|_2+\mu_{3n}\|\bb_3-\bb_3^*\|_2\big)\|\bb_1^{+}-\bb_1^*\|_2.
\end{eqnarray*}
Since  $\lambda_{1n}\geq 4\varepsilon_1+\big(\mu_{2n}\|\bb_2-\bb_2^*\|_2+\mu_{3n}\|\bb_3-\bb_3^*\|_2\big)/\Phi(\cS_1)$, we have
\begin{eqnarray}
\label{eqn:intermed_2}
\cP(\bb_1^+)-\cP(\bb_1^*) \leq \frac{1}{4}\cP(\bb_1^+-\bb_1^*)+\Phi(\cS_1)\|\bb_1^+-\bb_1^*\|_2.
\end{eqnarray}
Again, using the decomposability of $\cP$, the LHS of \eqref{eqn:intermed_2} can be decomposed by
\begin{eqnarray}\label{eqn:decom_pen}
  \cP(\bb_1^+)-\cP(\bb_1^*) &=& \cP(\bb_1^+-\bb_1^*+\bb_1^*)-\cP(\bb_1^*)\nonumber\\
  &=& \cP((\bb_1^+-\bb_1^*)_{\cS_1^{\bot}}+\bb_1^*+(\bb_1^+-\bb_1^*)_{\cS_1})-\cP(\bb_1^*)\nonumber\\
  &\geq& \cP((\bb_1^+-\bb_1^*)_{\cS_1^{\bot}})+\cP(\bb_1^*+(\bb_1^+-\bb_1^*)_{\cS_1})-\cP(\bb_1^*)\nonumber\\
  &\geq& \cP((\bb_1^+-\bb_1^*)_{\cS_1^{\bot}}) - \cP((\bb_1^+-\bb_1^*)_{\cS_1}),
\end{eqnarray}
where $\cS_1^{\perp}$ is the complement set of $\cS_1$. Equipped with \eqref{eqn:intermed_2},
\begin{eqnarray}
\label{eqn:cone}
3\cP((\bb_1^+-\bb_1^*)_{\cS_1^{\bot}})\leq 5\cP((\bb_1^+-\bb_1^*)_{\cS_1}) + 4\Phi(\cS_1)\|\bb_1^+-\bb_1^*\|_2.
\end{eqnarray}
By the definition of support space compatibility constant \eqref{def:support_constant},
$$ \cP((\bb_1^+-\bb_1^*)_{\cS_1})\leq \Phi(\cS_1)\|(\bb_1^+-\bb_1^*)_{\cS_1}\|_2\leq \Phi(\cS_1)\|\bb_1^+-\bb_1^*\|_2.$$
Together with $\cP(\bb_1^+-\bb_1^*)\leq \cP((\bb_1^+-\bb_1^*)_{\cS_1}) + \cP((\bb_1^+-\bb_1^*)_{\cS_1^{\bot}})$ and \eqref{eqn:cone}, we obtain
\begin{eqnarray}
\label{ineqn:decomposition_R}
\cP(\bb_1^+-\bb_1^*)\leq 4\Phi(\cS_1)\|\bb_1^+-\bb_1^*\|_2.
\end{eqnarray}

On the other hand, based on sparse strongly convex Condition \ref{con:strongly_convex},
\begin{eqnarray*}
  \cL(\bb_1^*,\bb_2, \bb_3)- \cL(\bb_1^+, \bb_2, \bb_3)-\langle\nabla_1\cL(\bb_1^*,\bb_2,\bb_3), \bb_1^*-\bb_1^+\rangle\leq -\frac{\gamma_{1n}}{2}\|\bb_1^+-\bb_1^*\|_2^2.
\end{eqnarray*}
with probability at least $1-\delta_2$. Plugging in \eqref{eqn:L}, we obtain with probability at least $1-(\delta_1+\delta_2+\delta_3),$
\begin{equation}
\label{ineqn:inter1}
\begin{split}
&\frac{\gamma_{1n}}{2}\|\bb_1^+-\bb_1^*\|_2^2\leq   \langle\nabla_1\cL(\bb_1^*,\bb_2,\bb_3), \bb_1^*-\bb_1^+\rangle + \cL(\bb_1^+,\bb_2, \bb_3)- \cL(\bb_1^*, \bb_2, \bb_3)\\
\leq & \ \varepsilon_1 \cP(\bb_1^+-\bb_1^*)+ \big(\mu_{2n}\|\bb_2-\bb_2^*\|_2+\mu_{3n}\|\bb_3-\bb_3^*\|_2\big)\|\bb_1^{+}-\bb_1^*\|_2 + \lambda_{1n}(\cP(\bb_1^*) - \cP(\bb_1^+)).
\end{split}
\end{equation}
 From \eqref{eqn:decom_pen}, 
\begin{eqnarray*}
\lambda_{1n}(\cP(\bb_1^*) - \cP(\bb_1^+)) &\leq& \lambda_{1n}\Big(\cP((\bb_1^+-\bb_1^*)_{\cS_1}) -\cP((\bb_1^+-\bb_1^*)_{\cS_1^{\bot}})\Big)\nonumber\\
&\leq& \lambda_{1n}\cP((\bb_1^+-\bb_1^*)_{\cS_1}).
\end{eqnarray*} 
Together with \eqref{ineqn:decomposition_R} and \eqref{ineqn:inter1},
\begin{eqnarray*}
\frac{\gamma_{1n}}{2}\|\bb_1^+-\bb_1^*\|_2^2 &\leq& \lambda_{1n} \Phi(\cS_1)\|\bb_1^+-\bb_1^*\|_2 + 4\varepsilon_1 \Phi(\cS_1)\|\bb_1^+-\bb_1^*\|_2\nonumber\\ &&+\big(\mu_{2n}\|\bb_2-\bb_2^*\|_2+\mu_{3n}\|\bb_3-\bb_3^*\|_2\big)\|\bb_1^{+}-\bb_1^*\|_2.
\end{eqnarray*}
Dividing by $\|\bb_1^{+}-\bb_1^*\|_2$ in both sides and plugging in the lower bound of $\lambda_{1n}$,  it yields that
\begin{eqnarray*}
  \|\bb_1^+-\bb_1^*\|_2&\leq& \frac{4\lambda_{1n}\Phi(\cS_1)}{\gamma_{1n}},
\end{eqnarray*}
 with probability at least $1-(\delta_1+\delta_2+\delta_3)$. This ends the proof.  \hfill $\blacksquare$\\  

Note that \eqref{eqn:generic_update} is a generic result since we have not provided a detail form for certain parameters.  Similar results also hold for the update of $\bb_2^{(t)}, \bb_3^{(t)}$ (see next corollary) and detailed proofs are omitted here.
\begin{corollary}\label{cor:one-step}
Suppose Conditions \ref{con:gradient_stability_b2}-\ref{con:stat_error_b3} hold, and the updates at time $t$ satisfy $\bb_1^{(t)}\in\cB_{\alpha_1, s_1}(\bb_1^*)$, $\bb_2^{(t)}\in\cB_{\alpha_2, s_2}(\bb_2^*)$, $\bb_3^{(t)}\in\cB_{\alpha_3, s_3}(\bb_3^*)$. With the regularization parameters $\lambda_{2n}^{(t)}, \lambda_{3n}^{(t)}$ satisfy
\begin{eqnarray*}
&&\lambda_{2n}^{(t)}\geq 4\varepsilon_2+(\mu_{1n}'\|\bb_1^{(t)}-\bb_1^*\|_2+\mu_{3n}'\|\bb_3^{(t)}-\bb_3^*\|_2)/\Phi(\cS_2)\\
&&\lambda_{3n}^{(t)}\geq 4\varepsilon_3+(\mu_{1n}^{''}\|\bb_1^{(t)}-\bb_2^*\|_2+\mu_{2n}^{''}\|\bb_2^{(t)}-\bb_3^*\|_2)/\Phi(\cS_3)
\end{eqnarray*}
and penalty $\cP$ fulfills the decomposable property, then the updates of $\bb_2,\bb_3$ at time $t+1$ satisfy 
\begin{eqnarray*}
&& \|\bb_2^{(t+1)}-\bb_2^*\|_2\leq 4\lambda_{2n}^{(t)}\Phi(\cS_2)/\gamma_{2n}\\
&& \|\bb_3^{(t+1)}-\bb_3^*\|_2\leq 4\lambda_{3n}^{(t)}\Phi(\cS_3)/\gamma_{3n},
\end{eqnarray*}
with probability at least $1-(\delta_1+\delta_2+\delta_3)$.
\end{corollary}

Now we are ready to prove the main theorem. Applying the result in Lemma \ref{lemma:one-step}, and plugging in the lower bound of $\lambda_{1n}^{(t)}$, we have 
\begin{eqnarray*}
  \|\bb_1^{(t+1)}-\bb_1^*\|_2\leq  \frac{4\mu_{2n}}{\gamma_{1n}}\|\bb_2^{(t)} - \bb_2^*\|_2 +\frac{4\mu_{3n}}{\gamma_{1n}} \|\bb_3^{(t)} - \bb_3^*\|_2+\frac{16\varepsilon_1\Phi(\cS_1)}{\gamma_{1n}}.
\end{eqnarray*}
Taking the square in both sides and noticing that $(a+b+c)^2 \leq 3(a^2+b^2+c^2)$,
\begin{eqnarray*}
  \|\bb_1^{(t+1)}-\bb_1^*\|_2^2\leq  3\Big(\frac{4\mu_{2n}}{\gamma_{1n}}\Big)^2\|\bb_2^{(t)} - \bb_2^*\|_2^2 +3\Big(\frac{4\mu_{3n}}{\gamma_{1n}}\Big)^2 \|\bb_3^{(t)} - \bb_3^*\|_2^2+3\Big(\frac{16\varepsilon_1\Phi(\cS_1)}{\gamma_{1n}}\Big)^2,
\end{eqnarray*}
with probability at least $1-(\delta_1+\delta_2+\delta_3)$. Similarly, applying Corollary \ref{cor:one-step}, we have
\begin{eqnarray*}
  &&\|\bb_2^{(t+1)}-\bb_2^*\|_2^2\leq  3\Big(\frac{4\mu'_{1n}}{\gamma_{2n}}\Big)^2\|\bb_1^{(t)} - \bb_1^*\|_2^2 +3\Big(\frac{4\mu'_{3n}}{\gamma_{2n}}\Big)^2 \|\bb_3^{(t)} - \bb_3^*\|_2^2+3\Big(\frac{16\varepsilon_2\Phi(\cS_2)}{\gamma_{2n}}\Big)^2\\
  &&\|\bb_3^{(t+1)}-\bb_3^*\|_2^2\leq  3\Big(\frac{4\mu^{''}_{1n}}{\gamma_{3n}}\Big)^2\|\bb_1^{(t)} - \bb_1^*\|_2^2 +3\Big(\frac{4\mu^{''}_{2n}}{\gamma_{3n}}\Big)^2 \|\bb_2^{(t)} - \bb_2^*\|_2^2+3\Big(\frac{16\varepsilon_3\Phi(\cS_3)}{\gamma_{3n}}\Big)^2,
\end{eqnarray*}
with probability at least $1-(\delta_1+\delta_2+\delta_3)$. Denote $\cE^{(t+1)} = \|\bb_1^{(t+1)}-\bb_1^*\|_2^2+\|\bb_2^{(t+1)}-\bb_2^*\|_2^2+\|\bb_3^{(t+1)}-\bb_3^*\|_2^2$. Adding the above three bounds together, it implies
\begin{eqnarray*}
  \cE^{(t+1)}&\leq&
  48\Big(\Big[\frac{\mu^{'2}_{1n}}{\gamma_{2n}^2}+\frac{\mu^{''2}_{1n}}{\gamma_{3n}^2}\Big]\|\bb_1^{(t)}-\bb_1^*\|_2^2 + \Big[\frac{\mu_{2n}^{2}}{\gamma_{1n}^{2}}+\frac{\mu_{2n}^{''2}}{\gamma_{3n}^2}\Big]\|\bb_2^{(t)}-\bb_2^*\|_2^2+\Big[\frac{\mu_{3n}^2}{\gamma_{1n}^2}+\frac{\mu_{3n}^{'2}}{\gamma_{2n}^2}\Big]\|\bb_3^{(t)}-\bb_3^*\|_2^2\Big)\\
  &&+768\Big(\frac{\varepsilon_1^2\Phi(\cS_1)^2}{\gamma_{1n}^2}+\frac{\varepsilon_2^2\Phi(\cS_2)^2}{\gamma_{2n}^2}+\frac{ \varepsilon_3^2\Phi(\cS_3)^2}{\gamma_{3n}^2}\Big).
\end{eqnarray*}
Define the contraction parameter
\begin{equation*}
\rho =288\max\{\mu_{1n}^{'2}, \mu_{1n}^{''2},\mu_{2n}^2, \mu_{2n}^{''2},\mu_{3n}^2, \mu_{3n}^{'2}\}/\min\{\gamma_{1n}^2, \gamma_{2n}^2, \gamma_{3n}^2\},
\end{equation*} then
\begin{eqnarray*}
  \cE^{(t+1)}\leq \rho\cE^{(t)}+ C_0\Big(\frac{\varepsilon_1^2\Phi(\cS_1)^2}{\gamma_{1n}^2}+\frac{\varepsilon_2^2\Phi(\cS_2)^2}{\gamma_{2n}^2}+\frac{ \varepsilon_3^2\Phi(\cS_3)^2}{\gamma_{3n}^2}\Big),
\end{eqnarray*}
with probability at least $1-3(\delta_1+\delta_2+\delta_3)$. This ends the proof. \hfill $\blacksquare$\\

\subsection{Proof of Theorem \ref{thm:star}}

Moreover, let $\alpha = \min\{\alpha_1, \alpha_2,\alpha_3\}$, $p = \max\{p_1, p_2, p_3\}$ and $s = \max\{s_1, s_2, s_3\}$, where $s_i$ is the cardinality of $\cS_i$ defined in \eqref{def:beta}.

Our proof consists of three steps. First, we verify Conditions \ref{con:gradient_stability}-\ref{con:strongly_convex} and \ref{con:stat_error} in Lemma \ref{lemma:rsc}-\ref{lemma:sc} for B-spline basis function and give explicit forms of Lipschitz-gradient parameter, sparse-strongly-convex parameter and statistical error. Second, we prove a generic contraction result by the induction argument. Last, we combine results in first two steps and achieve the final estimation rate.

At first, Lemma \ref{lemma:rsc} and \ref{lemma:oe} show that the loss function in \eqref{def:ERM} with B-spline basis function is sparse strongly convex and Lipschitz continuous. The proofs are deferred to Sections \ref{proof:rsc} and \ref{proof:oe}. 
\begin{lemma}\label{lemma:rsc}
  Consider $\{\psi_{jklh}(x)\}_{h=1}^{d_n}$ introduced in \eqref{eqn:addtive} are normalized B-spline basis functions and suppose Conditions Conditions \ref{con:nonpara_component}-\ref{con:noise} and \ref{con:signal} hold. When $\bb_1\in\cB_{\alpha, s}(\bb_1^*)$, $\bb_2\in \cB_{\alpha, s}(\bb_2^*)$, $\bb_3\in\cB_{\alpha, s}(\bb_3^*)$, the loss function $\cL(\cdot,\cdot,\cdot)$ is sparse strongly convex in its first argument, namely
    \begin{eqnarray}\label{ineqn:strongly_convex}
      \cL(\bb_1^*, \bb_2, \bb_3)-\cL(\bb_1, \bb_2, \bb_3)-\langle\nabla_1 \cL\big(\bb_1^*, \bb_2, \bb_3), \bb_1^*-\bb_1\rangle\leq -\frac{\gamma_{1n}}{2}\|\bb_1-\bb_1^*\|_2^2,
    \end{eqnarray}
    where  $\gamma_{1n}=C_1(1+o(1))Rd_n^{-1}s^2c_*^4$.
\end{lemma}
\begin{lemma}\label{lemma:oe}
  Suppose $\bb_2\in\cB_{\alpha,s}(\bb_2^*), \bb_3\in\cB_{\alpha,s}(\bb_3^*)$ and Conditions Conditions \ref{con:nonpara_component}-\ref{con:noise} and \ref{con:signal} hold. Considering the B-spline basis function, we have with probability at least $1-12/p$,
  \begin{eqnarray*}
    &&T_1 = \big\langle\nabla_1\tilde{\cL}(\bb_1^*, \bb_2, \bb_3^*)-\nabla_1\tilde{\cL}(\bb_1^*, \bb_2^*, \bb_3^*), \bb_1-\bb_1^* \big\rangle\leq \mu_{2n}\big\|\bb_1-\bb_1^*\big\|_2\big\|\bb_2-\bb_2^*\big\|_2\\
    &&T_2 = \big\langle\nabla_1\tilde{\cL}(\bb_1^*, \bb_2, \bb_3)-\nabla_1\tilde{\cL}(\bb_1^*, \bb_2, \bb_3^*), \bb_1-\bb_1^*\big\rangle\leq \mu_{3n}\big\|\bb_1-\bb_1^*\big\|_2\big\|\bb_3-\bb_3^*\big\|_2,
  \end{eqnarray*}
  where $\mu_{2n}=\mu_{3n} = 12(s^3R^2/d_n^2+C_0\sqrt{\log p/n})R^2s^2c^{*4}$.
\end{lemma}

The verification of Conditions \ref{con:strongly_convex_b2}-\ref{con:stat_error_b3} and derivation of $\gamma_{2n}, \gamma_{3n}, \mu_{1n}',\mu_{3n}^{'}, \mu_{1n}^{''}, \mu_{2n}^{''}$ remain the same and only differ in some constants. Thus, we let 
\begin{equation}\label{eqn:max_min}
\begin{split}
    & \max\{\mu_{2n}, \mu_{3n}, \mu_{1n}',\mu_{3n}^{'}, \mu_{1n}^{''}, \mu_{2n}^{''}\} = C_3(s^3R^2/d_n^2+\sqrt{\log p/n})R^2s^2c^{*4}\\
    & \min\{\gamma_{1n}, \gamma_{2n}, \gamma_{3n}\} = C_4(1+o(1))Rd_n^{-1}s^2c_*^4
    \end{split}
\end{equation}
for some absolute constant $C_3, C_4$.

Next lemma gives an explicit bound on statistical error for the update of $\bb_1$ when we utilize B-spline basis and choose the penalty $\cP$ to be group lasso penalty.
\begin{lemma}\label{lemma:sc}
  Suppose Conditions \ref{con:nonpara_component}-\ref{con:noise} and \ref{con:signal} hold and Consider $\{\psi_{jklh}(x)\}_{h=1}^{d_n}$ introduced in \eqref{eqn:addtive} to be normalized B-spline basis function. For $\bb_2\in\cB_{\alpha, s}(\bb_2^*)$, $\bb_3\in\cB_{\alpha,s}(\bb_3^*)$, we have with probability at least $1-C_0Rd_ns/n$, 
\begin{eqnarray*}
 &&\big\|\nabla_1\cL(\bb_1^*, \bb_2, \bb_3)-\nabla_1\tilde{\cL}(\bb_1^*, \bb_2, \bb_3)\big\|_{\cP^*}\\
 &\leq& C_1 Rc^{*4}\Big(\frac{s^5}{d_n^{\kappa-1/2}}\sqrt{\frac{\log (en)}{n}}+\frac{s^6}{d_n^{\kappa+1/2}}+\sigma \sqrt{\frac{s^4\log(pd_n)}{n}}\Big).
\end{eqnarray*}
for some absolute constants $C_0,C_1$, where $0<\kappa<1$ describes the smoothness of function class $\cH$ defined in \eqref{eqn:function_class}.
\end{lemma}

We complete the proof of Theorem \ref{thm:star} by the induction argument. When $t=1$, the initialization condition naturally holds by Condition \ref{con:initial}. Suppose $\|\bb_1^{(t)} - \bb_1^*\|_2\leq \alpha $, $\|\bb_2^{(t)} - \bb_2^*\|_2\leq \alpha $, $\|\bb_3^{(t)} - \bb_3^*\|_2\leq \alpha$ holds for some $t\geq 1$. For $t=t+1$, first we utilize the result in Lemma \ref{lemma:one-step} and plug in the lower bound of $\lambda_{1n}^{(t)}$,
\begin{eqnarray*}
  \|\bb_1^{(t+1)}-\bb_1^*\|_2&\leq&\frac{4\lambda_{1n}^{(t)}\Phi(\cS_1)}{\gamma_{1n}}\\
  &\leq& \frac{4\Phi(\cS_1)}{\gamma_{1n}}\Big(4\varepsilon_1+\big(\mu_{2n}\|\bb_2^{(t)}-\bb_2^*\|_2+\mu_{3n}\|\bb_3^{(t)}-\bb_3^*\|_2\big)/\Phi(\cS_1)\Big)\\
  &\leq&  \frac{16\Phi(\cS_1)\varepsilon_1}{\gamma_{1n}} + \frac{4\mu_{2n}}{\gamma_{1n}}\|\bb_2^{(t)}-\bb_2^*\|_2+\frac{4\mu_{3n}}{\gamma_{1n}}\|\bb_3^{(t)}-\bb_3^*\|_2\\
  &\leq& \frac{16\Phi(\cS_1)\varepsilon_1}{\gamma_{1n}} + \frac{4}{\gamma_{1n}}\Big(\mu_{2n}\alpha+\mu_{3n}\alpha\Big).
\end{eqnarray*}
As long as the statistical error $\varepsilon_1$ satisfies
\begin{equation}\label{eqn:stat_error1}
\varepsilon_1\leq \Big(1-\frac{4(\mu_{2n}+\mu_{3n})}{\gamma_{1n}}\Big)\frac{\alpha\gamma_{1n}}{40\Phi(\cS_1)},
\end{equation}
we have $\|\bb_1^{(t+1)}-\bb_1^*\|_2\leq \alpha$. The proofs for $\|\bb_2^{(t+1)}-\bb_2^*\|_2\leq \alpha$ and $\|\bb_3^{(t+1)}-\bb_3^*\|_2\leq \alpha$ are similar when $\varepsilon_2, \varepsilon_3$ satisfy
\begin{eqnarray}\label{eqn:stat_error2}
  &&\varepsilon_2\leq \Big(1-\frac{4(\mu_{1n}+\mu_{3n})}{\gamma_{2n}}\Big)\frac{\alpha\gamma_{2n}}{16\Phi(\cS_2)},\\
  &&\varepsilon_3\leq \Big(1-\frac{4(\mu_{1n}+\mu_{2n})}{\gamma_{3n}}\Big)\frac{\alpha\gamma_{3n}}{16\Phi(\cS_3)}.\nonumber
\end{eqnarray}

Second, when $\bb_2, \bb_3$ are fixed, the update scheme for $\bb_1$ exactly fits the one in \cite{huang2010variable} with group lasso penalty under B-spline basis function expansion. Define $\cS_1^{(t)} = \{j\in[p_1]|\|\bbeta_{1j}^{(t)}\|_2\neq 0\}$. Similar to the proof of first part in Theorem 1 in \cite{huang2010variable}, we could obtain $|\cS_1^{(t)}|\leq C_0|\cS_1|=C_0s$ for a finite constant $C_0>1$ with probability converging to 1. That means the number of non-zero elements in the estimator from group-lasso-type penalization is comparable with the size of true support. The guarantee for $\bb_2^{(t)},\bb_3^{(t)}$ remains the same. 

Therefore, we can conclude that $\bb_1^{(t)}\in\cB_{\alpha, s}(\bb_1^*), \bb_2^{(t)}\in \cB_{\alpha, s}(\bb_2^*),\bb_3^{(t)}\in \cB_{\alpha, s}(\bb_3^*)$ hold for any iteration $t=1,2,\ldots$ as long as the statistical error is sufficiently small such that \eqref{eqn:stat_error1}-\eqref{eqn:stat_error2} hold. We choose the tuning parameter $\lambda_{1n}^{(t)},\lambda_{2n}^{(t)}, \lambda_{3n}^{(t)}$ as defined in Condition \ref{con:regularization} with generic parameters specified in Lemma \ref{lemma:oe}, \ref{lemma:sc}. Repeatedly applying the result in Theorem \ref{thm:para} and summing from $t=1$ to $t=t+1$, one can provide a generic form of error updates,
\begin{eqnarray}\label{ineqn:cE}
  \cE^{(t+1)}\leq  \rho^{t+1}\cE^{(0)}+ \frac{1-\rho^{t+1}}{1-\rho}C_0\Big(\frac{\varepsilon_1^2\Phi(\cS_1)^2}{\gamma_{1n}^2}+\frac{\varepsilon_2^2\Phi(\cS_2)^2}{\gamma_{2n}^2}+\frac{ \varepsilon_3^2\Phi(\cS_3)^2}{\gamma_{3n}^2}\Big),
\end{eqnarray}
with probability at least $1-2(t+1)(\delta_1+\delta_2+\delta_3)$. As before, \eqref{ineqn:cE} still provides a generic form of error updates. 

Finally, we combine results from  Lemmas \ref{lemma:rsc}-\ref{lemma:sc}. According to \eqref{eqn:max_min}, the contraction parameter is upper bounded by
\begin{eqnarray*}
  \rho \leq \frac{288 C_3^2c^{*8}}{C_4^2c_*^8}\Big(\frac{s^6}{d_n^2}+\frac{d_n^2\log p }{n}\Big).
\end{eqnarray*}
When the sample size $n$ is large enough such that
\begin{equation}\label{eqn:sample_size}
d_n^2\geq \frac{1}{4}\frac{288C_3^2c^{*8}s^6}{C_4^2c_*^8}, \ n\geq \frac{1}{4}\frac{288C_3^2d_n^2(\log p)c^{*8}}{c_*^8C_4^2},
\end{equation}
one can guarantee $\rho\leq 1/2$. For group lasso penalty \eqref{eqn:group_lasso}, it has been shown in \cite{Wain2014} that $\max\{\Phi(\cS_1), \Phi(\cS_2), \Phi(\cS_3)\} = s$. Then, we can have an explicit form for the upper bound in \eqref{ineqn:cE},
\begin{eqnarray}\label{eqn:E}
  \cE^{(t+1)}&\leq & \rho^{t+1}\cE^{(0)}+ \frac{1-\rho^{t+1}}{1-\rho}3C_0 \frac{\max(\varepsilon_1^2, \varepsilon_2^2, \varepsilon_3^2)\max(\Phi(\cS_1)^2, \Phi(\cS_2)^2, \Phi(\cS_3)^2)}{\min(\gamma_{1n}^2, \gamma_{2n}^2, \gamma_{3n}^2)}\nonumber\\
  &\leq& \rho^{t+1}\cE^{(0)}+ \frac{1-\rho^{t+1}}{1-\rho}3\frac{C_0R^2}{(1+o(1))}\Big(\frac{s^2}{d_n^{-2}s^4c_*^8}\Big)c^{*8}3\Big(\frac{s^{10}}{d_n^{2\kappa-1}}\frac{\log ep}{n}+\frac{s^{12}}{d_n^{2\kappa+1}}+ \frac{\sigma^2\log (pd_n)}{n}\Big)\nonumber\\
  &=& \rho^{t+1}\cE^{(0)} + \frac{1-\rho^{t+1}}{1-\rho}\frac{9C_0c^{*8}R^2}{c_{*}^8(1+o(1))}\Big(
  \frac{s^8}{d_n^{2\kappa-3}}\frac{\log ep}{n}+\frac{s^{10}}{d_n^{2\kappa-1}}+\sigma^2\frac{d_n^2}{s^2}\frac{\log (pd_n)}{n}\Big),
\end{eqnarray}
with probability at least $1-C_0(t+1)(Rd_ns/n+1/p)$. From Conditions \ref{con:noise}-\ref{con:signal}, we known that $s, \sigma, c_*, c^*$ are all bounded by some absolute constants. Then \eqref{eqn:E} can be further simplified as 
\begin{equation*}
      \cE^{(t+1)}\leq \rho^{t+1}\cE^{(0)} + \frac{C_1R^2}{(1+o(1))}\frac{1-\rho^{t+1}}{1-\rho}\Big(
\frac{\log ep}{d_n^{2\kappa-3}n}+\frac{1}{d_n^{2\kappa-1}}+\sigma^2\frac{d_n^2\log (pd_n)}{s^2n}\Big).
\end{equation*}

To trade-off the statistical error part $(\sigma^2d_n^2\frac{\log pd_n}{s^2n})$ and approximation error part $(\frac{\log ep}{d_n^{2\kappa-3}n}+\frac{1}{d_n^{2\kappa-1}})$, one can take $d_n\asymp n^{\tfrac{1}{2\kappa+1}}$. Then the above bound will reduce to
\begin{equation*}
  \cE^{(t+1)}\leq\rho^{t+1}\cE^{(0)}+ \frac{C_1R^2}{(1-\rho)(1+o(1))}n^{-\tfrac{2\kappa-1}{2\kappa+1}}\log (pd_n),
\end{equation*}
with proper adjustments for the constant $C_1$. Moreover, when the total number of iterations is no smaller than
\begin{equation*}
T^*=\log\Big(\frac{(1-\rho)(1+o(1))}{C_1\cE^{(0)}}\frac{n^{\tfrac{2\kappa-1}{2\kappa+1}}}{\log (pd_n)}\Big) / \log(1/\rho),
\end{equation*}
we have with probability at least $1-C_0(T^*+1)(Rsn^{-\tfrac{2\kappa}{2\kappa+2}}+1/p)$,
\begin{equation*}
\cE^{(T^*)}\leq \frac{2C_1R^2}{(1-\rho)(1+o(1))}n^{-\tfrac{2\kappa-1}{2\kappa+1}}\log (pd_n),
\end{equation*}
as long as $n\geq C_2(\log p)^{\tfrac{2\kappa +1}{2\kappa -1}}$ for sufficiently large $C_2$. This sample complexity is sufficient to guarantee that \eqref{eqn:stat_error1}-\eqref{eqn:stat_error2} and \eqref{eqn:sample_size} hold under Conditions \ref{con:noise}-\ref{con:signal}. This ends the proof. \hfill $\blacksquare$\\

\subsection{Proof of Corollary \ref{cor:func}}

Recall that $\tilde{\cT}(\cX)=\sum_{h=1}^{d_n}\langle\cB_h^*,\cF_h(\cX)\rangle$, where $\cB_h^* = \sum_{r=1}^R\bbeta_{1hr}^*\circ \bbeta_{2hr}^*\circ \bbeta_{3hr}^*$, and $\hat{\cT}(\cX)=\sum_{h=1}^{d_n}\langle\hat{\cB}_h,\cF_h(\cX)\rangle$, where $\hat{\cB}_h = \sum_{r=1}^R\bbeta_{1hr}^{(T^*)}\circ \bbeta_{2hr}^{(T^*)}\circ \bbeta_{3hr}^{(T^*)}$. We make the following decomposition,
\begin{equation*}
\Big\|\hat{\cT}-\cT^*\Big\|_2^2 \leq 2\underbrace{\Big\|\hat{\cT}-\tilde{\cT}\Big\|_2^2}_{I_1}+2\underbrace{\Big\|\tilde{\cT}-\cT^*\Big\|_2^2}_{I_2},
\end{equation*}
where $\|f\|_2 = \sqrt{\int_{a}^bf^2(x)dP(x)}$.
Intuitively, $I_1$ quantifies the estimation error of $\{\cB_h^*\}_{h=1}^{d_n}$, while $I_2$ measures the overall approximation error by using B-spline basis function expansion. We bound $I_1$ and $I_2$ in two steps. 
\begin{enumerate}
    \item By the definition, it's easy to see
\begin{eqnarray*}
\sum_{h=1}^{d_n}\Big\|\hat{\cB}_h-\cB_h^*\Big\|_F^2\leq 3R\sum_{r=1}^R\sum_{h=1}^{d_n}\Big(\|\bbeta_{1hr}^{(T^*)}-\bbeta_{1hr}^*\|_2^2+ \|\bbeta_{2hr}^{(T^*)}-\bbeta_{2hr}^*\|_2^2+ \|\bbeta_{3hr}^{(T^*)}-\bbeta_{3hr}^*\|_2^2\Big).
\end{eqnarray*}
According to the basis property of spline expansions \citep{de1978practical}, we reach that
\begin{eqnarray*}
I_1&\leq& C_1 d_n^{-1}  \sum_{h=1}^{d_n}\Big\|\hat{\cB}_h-\cB_h^*\Big\|_F^2\\
&\leq & 3C_1 Rd_n^{-1}\sum_{r=1}^R\sum_{h=1}^{d_n}\Big(\|\bbeta_{1hr}^{(T^*)}-\bbeta_{1hr}^*\|_2^2+ \|\bbeta_{2hr}^{(T^*)}-\bbeta_{2hr}^*\|_2^2+ \|\bbeta_{3hr}^{(T^*)}-\bbeta_{3hr}^*\|_2^2\Big)\\
&=& 3C_1R d_n^{-1} \cE^{(T^*)}.
\end{eqnarray*}
According to Theorem \ref{thm:star}, 
\begin{equation}\label{eqn:I1_bound}
I_1\leq 3C_1R d_n^{-1} n^{-\tfrac{2\kappa-1}{2\kappa+1}}\log pd_n,
\end{equation}
with probability at least $1-C_0(T^*+1)(sn^{-\tfrac{2\kappa}{2\kappa+1}}+1/p)$.
\item By the assumption of CP-low-rankness, we have
\begin{eqnarray*}
I_2& =& \Big\|\sum_{h=1}^{d_n}\langle\cB_h^*,\cF_h(\cX)\rangle-\cT^*(\cX)\Big\|_2^2\\
&=&\Big\|\sum_{j=1}^{p_1}\sum_{k=1}^{p_2}\sum_{l=1}^{p_3}(f_{jkl}^{d_n}(\cX_{jkl})-f_{jkl}^*(\cX_{jkl}))\Big\|_2^2.
\end{eqnarray*}
According to Lemma \ref{lemma:spline_appro}, we have 
\begin{equation}\label{eqn:I2_bound}
I_2\leq C_2s^6d_n^{-2\kappa}.
\end{equation}
\end{enumerate}
Putting \eqref{eqn:I1_bound} and \eqref{eqn:I2_bound} together, we reach
\begin{equation*}
\Big\|\hat{\cT}-\cT^*\Big\|_2^2\leq 3C_1R d_n^{-1} n^{-\tfrac{2\kappa-1}{2\kappa+1}}\log pd_n + C_2s^6d_n^{-2\kappa}.
\end{equation*}
Note that under Condition \ref{con:nonpara_component}-\ref{con:initial}, both $R$ and $s$ are bounded. By taking $d_n\asymp n^{-\tfrac{1}{2\kappa+1}}$, we have
\begin{equation*}
\Big\|\hat{\cT}-\cT^*\Big\|_2^2= \cO_p\Big(n^{-\tfrac{2\kappa}{2\kappa+1}}\log pd_n\Big).
\end{equation*}
This ends the proof.
\hfill $\blacksquare$\\

\section*{Acknowledgments}

The authors thank the action editor Francis Bach and two reviewers for their helpful comments and suggestions which led to a much improved presentation. Will Wei Sun acknowledges support from the Office of Naval Research (ONR N00014-18-1-2759). Jingfei Zhang acknowledges support from the National Science Foundation (NSF DMS-2015190). Any opinions, findings, and conclusions or recommendations expressed in this material are those of the authors and do not necessarily reflect the views of the National Science Foundation, or the Office of Naval Research.

\bibliography{STAR}

\newpage
\renewcommand{\thesubsection}{\thesection.\arabic{subsection}}
\setcounter{equation}{0}
\setcounter{section}{0}  
\def\eop
{\hfill $\Box$
}

\begin{center}
{\Large\bf Supplementary Materials} \\
{\Large\bf Sparse Tensor Additive Regression} \\
\medskip
\large Botao Hao, Boxiang Wang, Pengyuan Wang,\\
Jingfei Zhang, Jian Yang, Will Wei Sun

\end{center}

In the supplementary, we present the definition and properties of the B-spline basis, some additional conditions for our theoretical results and the detailed proofs of Lemmas \ref{lemma:rsc}-\ref{lemma:sc}. 

\section{Properties of B-spline}\label{subsec:B-spline}

We formally define the $q$-th order B-splines with a set of $m$ internal knot sequences $k=\{0=k_0<k_1<\ldots<k_m<k_{m+1}=1\}$ recursively,
\begin{equation*}
  b_l^1(x)=
  \begin{cases}
    1, k_l\leq x<k_{l+1}\\
    0, \text{otherwise}
  \end{cases}
\end{equation*}
and 
\begin{equation}\label{def:B_spline}
    b_l^q(x)=\frac{x-k_l}{k_{l+q-1}-k_l}b_l^{q-1}(x)+\frac{k_{l+q}-x}{k_{l+q}-k_{l+1}}b_{l+1}^{q-1}.
\end{equation}
Then under some smoothness conditions, $f(x) \approx s(x)=\sum_lb_l^q(x)\beta_{l}=\bb(x)^{\top}\bbeta$, where $\bbeta_i\in\mathbb R^p$ with $p=m+q$. For the random variable $X$ satisfying Condition \ref{con:nonpara_component}, we have $\mathbb E[b_l^q(X)] \leq C_1 d_n^{-1}, \mathbb E[b_l^q(X)]^2 \leq C_2 d_n^{-1}$ for some constants $C_1$ and $C_2$. The detailed proofs refer to \cite{stone1985additive, huang2010variable, fan2011nonparametric}.

Additionally, we restate the result in \cite{huang2010variable} for the approximation error rate under B-spline basis function.
\begin{lemma}[\cite{stone1985additive,huang2010variable}]\label{lemma:spline_appro}
  Suppose Condition \ref{con:nonpara_component} holds and if the number of spline series is chosen by $d_n=\cO(n^{1/(2\kappa+1)})$. Then there exists an $f_{jkl}^{d_n}\in\cS_{n}$ satisfying
  \begin{eqnarray}
  \big\|f_{jkl}^{d_n}-f_{jkl}^*\big\|_2^2 = \cO_p(d_n^{-2\kappa}) = \cO_p(n^{-2\kappa/(2\kappa+1)}).
  \end{eqnarray}  
\end{lemma}

\section{Additional conditions for Section \ref{subsec:opt_stat}}\label{subsec:add_con}

In this section, we present addition conditions for Lipschitz-gradient (Conditions \ref{con:gradient_stability_b2}-\ref{con:gradient_stability_b3}), sparse strongly convex (Conditions \ref{con:strongly_convex_b2}-\ref{con:strongly_convex_b3}), and statistical error (Conditions \ref{con:stat_error_b2}-\ref{con:stat_error_b3}) for the update of $\bb_2$ and $\bb_3$. We define $\nabla_2\cL(\cdot,\cdot,\cdot)$ and $\nabla_3\cL(\cdot,\cdot,\cdot)$ are the gradient taken with respect to the second and the third argument.

\begin{condition}[\hspace*{-0.1cm}]\label{con:gradient_stability_b2} For $\bb_1\in\cB_{\alpha_1,s_1}(\bb_1^*), \bb_3\in\cB_{\alpha_3, s_3}(\bb_3^*)$, the noiseless gradient function $\nabla_2\tilde{\cL}(\cdot, \bb_2^*, \bb_3)$ satisfies $\mu_{1n}'$-Lipschitz-gradient condition, and $\nabla_2\tilde{\cL}(\bb_1^*, \bb_2^*, \cdot)$ satisfies $\mu_{3n}'$-Lipschitz-gradient condition. That is,
  \begin{eqnarray*}
  &&\big\langle\nabla_2\tilde{\cL}(\bb_1, \bb_2^*, \bb_3)-\nabla_2\tilde{\cL}(\bb_1^*, \bb_2^*, \bb_3), \bb_2-\bb_2^* \big\rangle\leq \mu_{1n}'\big\|\bb_2-\bb_2^*\big\|_2\big\|\bb_1-\bb_1^*\big\|_2\\
  &&\big\langle\nabla_2\tilde{\cL}(\bb_1^*, \bb_2^*, \bb_3)-\nabla_2\tilde{\cL}(\bb_1^*, \bb_2^*, \bb_3^*), \bb_2-\bb_2^*\big\rangle\leq \mu_{3n}'\big\|\bb_2-\bb_2^*\big\|_2\big\|\bb_3-\bb_3^*\big\|_2,
  \end{eqnarray*}
  with probability at least $1-\delta_1$.
\end{condition}

\begin{condition}[\hspace*{-0.1cm}]\label{con:gradient_stability_b3} For $\bb_1\in\cB_{\alpha_1,s_1}(\bb_1^*), \bb_2\in\cB_{\alpha_2, s_2}(\bb_2^*)$, the noiseless gradient function $\nabla_3\tilde{\cL}(\bb_1, \cdot, \bb_3^*)$ satisfies $\mu_{2n}^{''}$-Lipschitz-gradient condition, and $\nabla_3\tilde{\cL}(\cdot, \bb_2^*, \bb_3^*)$ satisfies $\mu_{1n}^{''}$-Lipschitz-gradient condition. That is,
  \begin{eqnarray*}
  &&\big\langle\nabla_3\tilde{\cL}(\bb_1, \bb_2, \bb_3^*)-\nabla_3\tilde{\cL}(\bb_1, \bb_2^*, \bb_3^*), \bb_3-\bb_3^* \big\rangle\leq \mu_{2n}^{''}\big\|\bb_3-\bb_3^*\big\|_2\big\|\bb_2-\bb_2^*\big\|_2\\
  &&\big\langle\nabla_3\tilde{\cL}(\bb_1, \bb_2^*, \bb_3^*)-\nabla_3\tilde{\cL}(\bb_1^*, \bb_2^*, \bb_3^*), \bb_3-\bb_3^*\big\rangle\leq \mu_{1n}^{''}\big\|\bb_3-\bb_3^*\big\|_2\big\|\bb_1-\bb_1^*\big\|_2,
  \end{eqnarray*}
  with probability at least $1-\delta_1$.
\end{condition}

\begin{condition}[\hspace*{-0.1cm}]\label{con:strongly_convex_b2}
  For any $\bb_1\in\cB_{\alpha_1,s_1}(\bb_1^*), \bb_3\in\cB_{\alpha_3,s_3}(\bb_3^*)$, the loss function $\cL(\cdot,\cdot,\cdot)$ is sparse strongly convex in its first variable, namely
  \begin{eqnarray*}
    \cL(\bb_1, \bb_2^*, \bb_3)-\cL(\bb_1, \bb_2, \bb_3)-\langle\nabla_2 \cL\big(\bb_1, \bb_2^*, \bb_3), \bb_2^*-\bb_2\rangle\geq \frac{\gamma_{2n}}{2}\|\bb_2-\bb_2^*\|_2^2,
  \end{eqnarray*}
  with probability at least $1-\delta_2$. Here, $\gamma_{2n}>0$ is the strongly convex parameter.
\end{condition}

\begin{condition}[\hspace*{-0.1cm}]\label{con:strongly_convex_b3}
  For any $\bb_1\in\cB_{\alpha_1,s_1}(\bb_1^*), \bb_2\in\cB_{\alpha_2,s_2}(\bb_2^*)$, the loss function $\cL(\cdot,\cdot,\cdot)$ is sparse strongly convex in its first variable, namely
  \begin{eqnarray*}
    \cL(\bb_1, \bb_2, \bb_3^*)-\cL(\bb_1, \bb_2, \bb_3)-\langle\nabla_3 \cL\big(\bb_1, \bb_2, \bb_3^*), \bb_3*-\bb_3\rangle\geq \frac{\gamma_{3n}}{2}\|\bb_3-\bb_3^*\|_2^2,
  \end{eqnarray*}
  with probability at least $1-\delta_2$. Here, $\gamma_{3n}>0$ is the strongly convex parameter.
\end{condition}

\begin{condition}[\hspace*{-0.1cm}]\label{con:stat_error_b2}
  For any $\bb_1\in \cB_{\alpha_1,s_1}(\bb_1^*)$, $\bb_3\in\cB_{\alpha_3,s_3}(\bb_3^*)$, we have with probability at least $1-\delta_3$,
  \begin{equation*}
  \big\|\nabla_2\cL(\bb_1, \bb_2^*, \bb_3)-\nabla_2\tilde{\cL}(\bb_1, \bb_2^*, \bb_3)\big\|_{\cP^*}\leq \varepsilon_2.
  \end{equation*}
\end{condition}

\begin{condition}[\hspace*{-0.1cm}]\label{con:stat_error_b3}
  For any $\bb_1\in \cB_{\alpha_1,s_1}(\bb_1^*)$, $\bb_2\in\cB_{\alpha_2,s_2}(\bb_2^*)$, we have with probability at least $1-\delta_3$,
  \begin{equation*}
  \big\|\nabla_3\cL(\bb_1, \bb_2, \bb_3^*)-3\tilde{\cL}(\bb_1, \bb_2, \bb_3^*)\big\|_{\cP^*}\leq \varepsilon_3.
  \end{equation*}
\end{condition}

\section{Proofs of Lemmas \ref{lemma:rsc}-\ref{lemma:sc}}
In this section, we present the proof of Lemmas \ref{lemma:rsc}-\ref{lemma:sc}. If $X$ is sub-Gaussian random variable, then its $\phi_2$-Orlicz norm can be bounded such that $\|X\|_{\phi_2}\leq C_1$ for some absolute constant. If $X$ is sub-exponential random variable, then its $\phi_1$-Orlicz norm can be bounded such that $\|X\|_{\phi_1}\leq C_2$ for some absolute constant $C_2$. 
\subsection{Proof of Lemma \ref{lemma:rsc}}\label{proof:rsc}

Recall that $\bb_1=(\vartheta_{11}^{\top}, \ldots, \vartheta_{1p_1}^{\top})^{\top}\in\mathbb R^{Rd_np_1\times 1}$. Define $\cS_1'=\{j\in[p]|\|\vartheta_{1j}\|_2\neq 0 \cup \|\vartheta_{1j}^*\|_2\neq 0\}$, $\bF_{\cS_1'}^1=(\bF_j^1\in\mathbb R^{n\times Rd_n},j\in \cS_1')$, $\bb_{1\cS_1'}=(\bbeta_{1j} \in\mathbb R^{Rd_n\times 1},j\in \cS_1')$. Since $\bb_1\in \cB_{\alpha, s}(\bb_1^*)$, we know that $|\cS_1'|=C_0 s$ for some positive constant $C_0\geq 1$ not depending on $s$. Without loss of generality, assume $|\cS_1'|=\{1,\cdots,C_0s\}$. First, we do some simplifications for the left side of \eqref{ineqn:strongly_convex}. According to the derivation in \eqref{def:gradient}, we have
\begin{eqnarray*}
&&\cL(\bb_1^*, \bb_2, \bb_3)-\cL(\bb_1, \bb_2, \bb_3)\\
&=& \frac{1}{n}\Big(\bb_1^{*\top}\bF^{1\top}\bF^1\bb_1^*- \bb_1^{\top}\bF^{1\top}\bF^1\bb_1-2\by^{\top}\bF^1\bb_1^*+2\by^{\top}\bF^1\bb_1\Big)\\
&=& \frac{1}{n}\Big(\bb_{1\cS_1'}^{*\top}\bF^{1\top}_{\cS_1'}\bF^1_{\cS_1'}\bb_{1\cS_1'}^*- \bb_{1\cS_1'}^{\top}\bF^{1\top}_{\cS_1'}\bF^1_{\cS_1'}\bb_{1\cS_1'}-2\by^{\top}\bF^1_{\cS_1'}\bb_{1\cS_1'}^*+2\by^{\top}\bF^1_{\cS_1'}\bb_{1\cS_1'}\Big),
\end{eqnarray*}
and 
\begin{eqnarray*}
&&\langle\nabla_1 \cL\big(\bb_1^*, \bb_2, \bb_3), \bb_1^*-\bb_1\rangle \\
&=& \frac{2}{n}\Big(\bb_1^{*\top}\bF^{1\top}\bF^1\bb_1^*- \bb_1^{*\top}\bF^{1\top}\bF^1\bb_1-\by^{\top}\bF^1\bb_1^*+\by^{\top}\bF^1\bb_1\Big)\\
&=& \frac{2}{n}\Big(\bb_{1\cS_1'}^{*\top}\bF^{1\top}_{\cS_1'}\bF^1_{\cS_1'}\bb_{1\cS_1'}^*- \bb_{1\cS_1'}^{*\top}\bF^{1\top}_{\cS_1'}\bF^1_{\cS_1'}\bb_{1\cS_1'}-\by^{\top}\bF^1_{\cS_1'}\bb_{1\cS_1'}^*+\by^{\top}\bF^1_{\cS_1'}\bb_{1\cS_1'}\Big).
\end{eqnarray*}
Putting the above two equations together, we reach
\begin{eqnarray*}
  &&\cL(\bb_1^*, \bb_2, \bb_3)-\cL(\bb_1, \bb_2, \bb_3)-\langle\nabla_1 \cL\big(\bb_1^*, \bb_2, \bb_3), \bb_1^*-\bb_1\rangle\\
  &=& (\bb_{1\cS_1'}-\bb_{1\cS_1'}^*)^{\top}\Big(-\frac{\bF^{1\top}_{\cS_1'}\bF^1_{\cS_1'}}{n}\Big)(\bb_{1\cS_1'}-\bb_{1\cS_1'}^*).
\end{eqnarray*}
It remains to prove
\begin{equation*}
(\bb_{1S_1}-\bb_{1S_1}^*)^{\top}\Big(\frac{\bF_{S_1}^{1\top}\bF_{S_1}^1}{n}\Big)(\bb_{1S_1}-\bb_{1S_1}^*)\geq  \frac{\gamma_{1n}}{2}\|\bb_{1S_1}-\bb_{1S_1}^*\|_2^2.
\end{equation*}
If one can show that $\bF_{S_1}^{1\top}\bF_{S_1}^1/n\succeq \tilde{m} \bI_{C_0s}$ i.e. the minimal eigenvalue $\sigma_{\min}(\bF_{S_1}^{1\top}\bF_{S_1}^1/n)\geq \tilde{m} $ for some positive $\tilde{m} \in\mathbb R$, then we have the strongly convex parameter $\gamma_{1n} =\tilde{m} $. Let $\ba=(\ba_1^{\top}, \ldots, \ba_{C_0s}^{\top})^{\top}$ where $\ba_j = (\ba_{j1}^{\top},\ldots, \ba_{jR}^{\top})^{\top}\in\mathbb R^{Rd_n\times 1}$ and $\ba_{jr}\in\mathbb R^{d_n\times 1}$. Our proof consists of two steps. 

\textbf{Step One.} Consider a single coordinate $\bF_j^1$. For $k\in[p]$ and $j\in[p]$, define
\begin{gather*}
Z_{jkl}=
\begin{pmatrix} 
\psi_{jkl1}([\cX_1]_{jkl}) &\cdots&\psi_{jkld_n}([\cX_1]_{jkl}) \\ 
\vdots&\ddots&\vdots\\
\psi_{jkl1}([\cX_n]_{jkl})&\cdots&\psi_{jkld_n}([\cX_n]_{jkl})
\end{pmatrix}\in \mathbb R^{n\times d_n},
\end{gather*}

\begin{gather*}
\bD_{klr}=
\begin{pmatrix} 
\beta_{21rk}\beta_{31rl} & & \\ 
&\ddots&\\
& &\beta_{2d_nrk}\beta_{3d_nrl}
\end{pmatrix}\in\mathbb R^{d_n\times d_n}.
\end{gather*}
By using the triangle inequality and Lemma 3 in \cite{stone1985additive}, we have for $j\in[C_0s]$,
\begin{eqnarray}
C_1\sum_{k=1}^{p}\sum_{l=1}^{p}\sum_{r=1}^R\Big\|Z_{jkl}\bD_{klr}\ba_{jr}\Big\|_2^2\leq \|\bF_j^1\ba_j\|_2^2,
\end{eqnarray}
 where $C_1$ is some positive constant. Divided by $n$ in both sides, we have
\begin{eqnarray*}
  C_1\sum_{k=1}^{p}\sum_{l=1}^{p}\sum_{r=1}^R\ba_{jr}^{\top}\bD_{klr}^{\top}\frac{Z_{jkl}^{\top}Z_{jkl}}{n}\bD_{klr}\ba_{jr}\leq \ba^{\top}_j\frac{\bF_j^{1\top}\bF_j^1}{n}\ba_j.
\end{eqnarray*}
According to Lemma 6.2 in \cite{zhou1998local}, there exists certain constants $C_2$ and $C_3$ such that
\begin{eqnarray}
C_2(1+o(1))d_n^{-1}\leq \sigma_{\min}\Big(\frac{Z_{jkl}^{\top}Z_{jkl}}{n}\Big)\leq \sigma_{\max}\Big(\frac{Z_{jkl}^{\top}Z_{jkl}}{n}\Big)\leq C_3(1+o(1))d_n^{-1}.
\end{eqnarray}
holds for any $k,l$. Since $\sigma_{\min}(\bA\bB)\geq \sigma_{\min}(\bA)\sigma_{\min}(\bB)$, we can bound the minimum eigenvalue of the weighted B-spline design matrix,
\begin{eqnarray*}
  \sigma_{\min}\Big(\bD_{klr}^{\top}\frac{Z_{jkl}^{\top}Z_{jkl}}{n}\bD_{klr}\Big)\geq C_2(1+o(1))d_n^{-1} (\min_{h} \beta_{2hk}\beta_{3hl})^2.
\end{eqnarray*}
This will enable us to bound the smallest eigenvalue of $\bF_j^{1\top}\bF_j^1/n$ as follows,
\begin{eqnarray*}
  \ba_j^{\top}\frac{\bF_j^{1\top}\bF_j/n}{\|\ba_j\|_2^2}\ba_j&\geq&C_1\sum_{k=1}^{p}\sum_{l=1}^{p}\sum_{r=1}^R\ba_{jr}^{\top}\frac{(\bD_{klr}^{\top}Z_{jkl}^{\top}Z_{jkl}\bD_{klr})/n}{\|\ba_{jr}\|_2^2}\ba_{jr}\\
  &\geq&  C_1C_2(1+o(1))d_n^{-1}\sum_{r=1}^R\min_h\big(\sum_{k=1}^p\beta_{2hrk}^{*2}-\alpha^2\big)\min_h\big(\sum_{l=1}^p\beta_{3hrl}^{*2}-\alpha^2\big)\\
        &\geq&  C_1C_2(1+o(1))Rd_n^{-1}(sc_*^2-\alpha^2)^2\\
  &\geq& \frac{1}{4}C_1C_2(1+o(1))Rd_n^{-1}s^2c_*^4,
\end{eqnarray*}
where last inequality is due to $\bb_2\in \cB_{\alpha, s}(\bb_2^*)$, $\bb_3\in \cB_{\alpha, s}(\bb_3^*)$ for $\alpha\leq c_*\sqrt{s/2}$ and Condition \ref{con:signal}. Let $C_1 = C_1/4$.  Therefore, for every $j\in[C_0s]$,
\begin{eqnarray}
\label{ineqn:bounded_eigenvalue}
\sigma_{\min}\Big(\frac{\bF_j^{1\top}\bF_j^1}{n}\Big)\geq C_1C_2(1+o(1))Rd_n^{-1}s^2c_*^4.
\end{eqnarray}

\textbf{Step Two.} By the triangle inequality, 
\begin{eqnarray*}
  C_4 \Big(\sum_{j=1}^{C_0s}\|\bF_j^1\bb_j\|_2^2\Big)\leq \|\bF_{S_1}^1\bb\|_2^2 = \bb^{\top}\bF_{S_1}^{1\top}\bF_{S_1}^1 \bb,
\end{eqnarray*}
for some constant $C_4$, which implies
\begin{eqnarray*}
  \ba^{\top}\frac{\bF_{S_1}^{1\top}\bF_{S_1}^1/n}{\|\ba\|_2^2}\ba\geq C_4\Big(\frac{\sum_{j=1}^{C_0s}\|\bF_j^1\ba_j\|_2^2}{n\|\ba\|_2^2}\Big).\end{eqnarray*}
Together with \eqref{ineqn:bounded_eigenvalue}, we have
\begin{eqnarray*}
\ba^{\top}\frac{\bF_{S_1}^{1\top}\bF_{S_1}^1/n}{\|\ba\|_2^2}\ba\geq C_1C_2C_4(1+o(1))d_n^{-1} s^2c_*^4
\end{eqnarray*}
holds for any $\ba$. Setting $C_1 = C_1C_2C_4$, it essentially implies
\begin{eqnarray*}
\sigma_{\min}(\frac{1}{n}\bF_{S_1}^{1\top}\bF_{S_1})\geq C_1(1+o(1))Rd_n^{-1}s^2c_*^4,
\end{eqnarray*}
for some constant $C_1$. We can say the sparse strong convexity holds with $\gamma_{1n}=C_1(1+o(1))Rd_n^{-1} s^2c_*^4$. \hfill $\blacksquare$\\

\subsection{Proof of Lemma \ref{lemma:oe}}\label{proof:oe}

 For notation simplicity, we denote 
$$
g_i(\bb_1,\bb_2, \bb_3)=\sum_{h=1}^{d_n}\langle\cF_h(\cX_i), \sum_{r=1}^R\bbeta_{1hr}\circ\bbeta_{2hr}\circ\bbeta_{3hr}\rangle,
$$
where $\cF_h(\cX_i)$ is defined in \eqref{eqn:F_n}.
According to the definition of the gradient function, we can rewrite the following inner product as
\begin{eqnarray}
&&\big\langle\nabla_1\tilde{\cL}(\bb_1, \bb_2, \bb_3), \bb_1^+-\bb_1^*\big\rangle\nonumber \\
&=& \frac{2}{n}\sum_{i=1}^n\Big[g_i(\bb_1,\bb_2,\bb_3)g_i(\bb_1^+-\bb_1^*, \bb_2,\bb_3)-g_i(\bb_1^*, \bb_2^*, \bb_3^*)g_i(\bb_1^+-\bb_1^*, \bb_2,\bb_3)\Big].
\end{eqnarray}
We will bound $T_1$ first. The bound for $T_2$ remains similar. Let's decompose $T_1$ by three parts,
\begin{eqnarray*}
T_1 &=& \big\langle\nabla_1\tilde{\cL}(\bb_1^*, \bb_2, \bb_3^*)-\nabla_1\tilde{\cL}(\bb_1^*, \bb_2^*, \bb_3^*), \bb_1-\bb_1^* \big\rangle\\
&=& \frac{2}{n}\sum_{i=1}^n\Big[g_i(\bb_1^*,\bb_2,\bb_3)g_i(\bb_1^+-\bb_1^*, \bb_2,\bb_3)-g_i(\bb_1^*, \bb_2^*, \bb_3^*)g_i(\bb_1^+-\bb_1^*, \bb_2,\bb_3)\\
&& -g_i(\bb_1^*,\bb_2^*,\bb_3)g_i(\bb_1^+-\bb_1^*, \bb_2^*,\bb_3)+g_i(\bb_1^*, \bb_2^*, \bb_3^*)g_i(\bb_1^+-\bb_1^*, \bb_2^*,\bb_3)\Big]\\
&=&\underbrace{\frac{2}{n}\sum_{i=1}^n\Big[g_i(\bb_1^*,\bb_2-\bb_2^*,\bb_3)g_i(\bb_1^+-\bb_1^*,\bb_2,\bb_3)\Big]}_{T_{11}}\\
&&+\underbrace{\frac{2}{n}\sum_{i=1}^n\Big[g_i(\bb_1^*,\bb_2^*,\bb_3)g_i(\bb_1^+-\bb_1^*,\bb_2-\bb_2^*,\bb_3)\Big]}_{T_{12}}\\
&&-\underbrace{\frac{2}{n}\sum_{i=1}^n\Big[g_i(\bb_1^*,\bb_2^*,\bb_3^*)g_i(\bb_1^+-\bb_1^*,\bb_2-\bb_2^*,\bb_3)\Big]}_{T_{13}}.
\end{eqnarray*}
By writing explicitly of $g_i(\bb_1,\bb_2,\bb_3)$,  
\begin{eqnarray*}
g_i(\bb_1,\bb_2,\bb_3) &=& \sum_{h=1}^{d_n}\Big(\sum_{j=1}^p\sum_{k=1}^p\sum_{l=1}^p[\cF_h(\cX_i)_{jkl}]\sum_{r=1}^R\beta_{1hrj}\beta_{2hrk}\beta_{3hrl}\Big)\\
&=& \sum_{h=1}^{d_n}\Big(\sum_{j=1}^p\sum_{k=1}^p\sum_{l=1}^p\psi_{jklh}([\cX_i]_{jkl})\sum_{r=1}^R\beta_{1hrj}\beta_{2hrk}\beta_{3hrl}\Big).
\end{eqnarray*}
Since $\sup_x|\psi_{jklh}(x)|\leq 1$, the $\phi_2$-Orlicz norm for each individual component can be bounded by
\begin{eqnarray*}
\Big\|\psi_{jklh}([\cX_i]_{jkl})\sum_{r=1}^R\beta_{1hrj}\beta_{2hrk}\beta_{3hrl}\Big\|_{\phi_2}\leq |\sum_{r=1}^R\beta_{1hrj}\beta_{2hrk}\beta_{3hrl}|, \ \text{for} \ j,k,l\in[p].
\end{eqnarray*}
Based on rotation invariance, we obtain 
\begin{eqnarray*}
\Big\|g_i(\bb_1,\bb_2,\bb_3)\Big\|_{\phi_2} \leq \Big(\sum_{h=1}^{d_n}\sum_{j=1}^p\sum_{k=1}^p\sum_{l=1}^p(\sum_{r=1}^R\beta_{1hrj}\beta_{2hrk}\beta_{3hrl})^2\Big)^{\tfrac{1}{2}}.
\end{eqnarray*}

In the following, we will bound the expectation of $g_i(\bb_1^*,\bb_2-\bb_2^*,\bb_3)g_i(\bb_1^+-\bb_1^*,\bb_2,\bb_3)$. By the property of B-spline basis function (See Section \ref{subsec:B-spline}) and Cathy-Schwarz inequality,
\begin{eqnarray*}
\mathbb E\big(g_i(\bb_1,\bb_2,\bb_3)\big) &=& \sum_{h=1}^{d_n}\Big(\sum_{j=1}^p\sum_{k=1}^p\sum_{l=1}^p\mathbb E[\cF_h(\cX_i)_{jkl}]\sum_{r=1}^R\beta_{1hrj}\beta_{2hrk}\beta_{3hrl}\Big)\\
&\leq& \frac{1}{d_n}\sum_{h=1}^{d_n}\sum_{j=1}^p\sum_{k=1}^p\sum_{l=1}^p\sum_{r=1}^R\beta_{1hrj}\beta_{2hrk}\beta_{3hrl}\\
&\leq&\frac{s^{\tfrac{3}{2}}R}{d_n}\Big(\sum_{h=1}^{d_n}\sum_{j=1}^p\sum_{k=1}^p\sum_{l=1}^p\sum_{r=1}^R\beta_{1hrj}^2\beta_{2hrk}^2\beta_{3hrl}^2\Big)^{\tfrac{1}{2}}.
\end{eqnarray*}
  Combining the above ingredients together with Hoeffding's inequality (See Lemma \ref{lemma:hoffding}), we obtain with probability at least $1-1/p$,
  \begin{eqnarray*}
  T_{11}&\leq& 2\Big[\frac{s^3R^2}{d_n^2}+C_0\sqrt{\frac{\log p}{n}}\Big]\Big(\sum_{h=1}^{d_n}\sum_{j=1}^p\sum_{k=1}^p\sum_{l=1}^p\sum_{r=1}^R\beta_{1hrj}^{*2}(\beta_{2hrk}-\beta_{2hrk}^*)^2\beta_{3hrl}^2\Big)^{\tfrac{1}{2}}\\
  &&\times \Big(\sum_{h=1}^{d_n}\sum_{j=1}^p\sum_{k=1}^p\sum_{l=1}^p\sum_{r=1}^R(\beta_{1hrj}^{+}-\beta_{1hrj}^{*})\beta_{2hrk}^2\beta_{3hrl}^2\Big)^{\tfrac{1}{2}}.
  \end{eqnarray*}
   Noting that $\bb_2\in \cB_{\alpha, s}(\bb_2^*), \bb_3\in \cB_{\alpha, s}(\bb_3^*)$, we have
  \begin{eqnarray*}
  T_{11}&\leq& 2\Big[\frac{s^3R^2}{d_n^2}+C_0\sqrt{\frac{\log p}{n}}\Big]\max_h(\sum_{j=1}^p\beta_{1hrj}^{*2})^{\tfrac{1}{2}}\max_h(\sum_{k=1}^p\beta_{2hrk}^{*2})^{\tfrac{1}{2}}\max_h(\sum_{l=1}^p\beta_{3hrl}^{*2})\big\|\bb_2-\bb_2^*\big\|_2\big\|\bb_1^+-\bb_1^*\big\|_2\\
  &\leq&2\Big[\frac{s^3R^2}{d_n^2}+C_0\sqrt{\frac{\log p}{n}}\Big]s^2R^2c^{*4}\big\|\bb_2-\bb_2^*\big\|_2\big\|\bb_1^+-\bb_1^*\big\|_2,
  \end{eqnarray*}
where the last inequality is from Condition \ref{con:signal}. The upper bounds for $T_{12}$ and $T_{13}$ are similar. Putting them together, with probability at least $1-6/p$,
  \begin{eqnarray*}
  |T_1|\leq 6\Big[\frac{s^3R^2}{d_n^2}+C_0\sqrt{\frac{\log p}{n}}\Big]R^2s^2c^{*4}\|\bb_2-\bb_2^*\|_2\|\bb_1^+-\bb_1^*\|_2.
  \end{eqnarray*}
Similarly, we can get the bound for $T_2$. This ends the proof. \hfill $\blacksquare$\\

\subsection{Proof of Lemma \ref{lemma:sc}}\label{proof:sc}

 Recall that $\cP^*$ is the dual norm of group lasso penalty $\cP$. With a little abuse of notations, we define $\bepsilon=(\epsilon_1,\ldots,\epsilon_n)^{\top}$ and $\cT^*(\cX)=(\cT^*(\cX_1),\ldots,\cT^*(\cX_n))^{\top}$ in this section. According to the derivation of the gradient function in \eqref{def:gradient}, we decompose the error by an spline approximation error term ($T_1$) and a statistical term ($T_2$) as follows,
\begin{eqnarray*}
&&\Big\|\nabla_1\cL(\bb_1^*, \bb_2, \bb_3)-\nabla_1\tilde{\cL}(\bb_1^*, \bb_2, \bb_3)\Big\|_{\cP^*}\\
&=& \Big\|\frac{2}{n}\bF^{1\top}(\bF^1\bb_1^*-\by)-\frac{2}{n}\bF^{1\top}(\bF^1\bb_1^*-\bF^{1*}\bb_1^*)\Big\|_{\cP^*}\\
&=& \Big\|\frac{2}{n}\bF^{1\top}(\by-\bF^{1*}\bb_1^*)\Big\|_{\cP^*} =\Big\|\frac{2}{n}\bF^{1\top}(\cT^*(\cX)-\bF^{1*}\bb_1^*+\bepsilon)\Big\|_{\cP^*}\\
&\leq&  \underbrace{\Big\|\frac{2}{n}\bF^{1\top}(\cT^*(\cX)-\bF^{1*}\bb_1^*)\Big\|_{\cP^*}}_{T_1}+\underbrace{\Big\|\frac{2}{n}\bF^{1\top}\bepsilon\Big\|_{\cP^*}}_{T_2}.
\end{eqnarray*}

\textbf{Step One: Bounding $T_1$.} Denote $A_1 = \{j\in [p]|\|\bF^1_{j}\|_2\neq 0\}$. Since  $\bb_2\in \cB_{\alpha, s}(\bb_2^*)$, $\bb_3\in \cB_{\alpha, s}(\bb_3^*)$, it's easy to see $|A_1|\leq C_0 s$ for some constant $C_0$ not depending on $n,p,s$. By the definition of dual norm $\cP^*$ (See Definition \ref{def:dual_norm}), we obtain
\begin{eqnarray}\label{ineqn:T1}
T_1&=&\Big\|\frac{2}{n}\sum_{i=1}^n\bF_i^1\Big(\cT^*(\cX_i)-[\bF^{1*}\bb_1^*]_i\Big)\Big\|_{\cP^*}\nonumber\\
&=& \max_{j\in A_1}\Big\|\frac{2}{n}\sum_{i=1}^n\bF_{ij}^1\Big(\cT^*(\cX_i)-[\bF^{1*}\bb_1^*]_i\Big)\Big\|_2\nonumber\\
&\leq& \max_{i\in[n]}\Big| \cT^*(\cX_i)-[\bF^{1*}\bb_1^*]_i\Big|\max_{j\in A_1}\Big\|\frac{2}{n}\sum_{i=1}^n\bF_{ij}^1\Big\|_2.
\end{eqnarray}
Note that the first part of \eqref{ineqn:T1} fully comes from the approximation error using B-spline basis functions for the nonparametric component. We bound $T_1$ in three steps as follows.
\begin{enumerate}
  \item  To bound the first part, we use Lemma \ref{lemma:spline_appro} which quantifies the approximation error for a single component. To ses this, there exists a positive constant $C_1$ such that
   $$
   \Big|f_{jkl}^{d_n}([\cX_i]_{jkl})-f_{jkl}^*([\cX_i]_{jkl})\Big| \leq C_1 d_n^{-\kappa}, \ j,k,l\in[p].
   $$ 
   For the whole nonparametric function $\cT^*$, we utilize the CP-low-rankness assumption \eqref{eqn:cp_dec} and group sparse assumption \eqref{def:beta}, which indicates
  \begin{equation}\label{eqn:spline_error}
  \max_{i\in[n]}\Big| \cT^*(\cX_i)-[\bF^{1*}\bb_1^*]_i\Big|=\max_{i\in[n]}\Big|\sum_{j=1}^p\sum_{k=1}^p\sum_{l=1}^p\Big(f_{jkl}^{d_n}([\cX_i]_{jkl})-f_{jkl}^*([\cX_i]_{jkl})\Big)\Big| \leq C_1 s^3 d_n^{-\kappa}.
  \end{equation}
   \item To bound the second part, by the definition of $\bF^1_{ij}$, we have
   \begin{equation*}
   \begin{split}
        \frac{1}{n}\sum_{i=1}^n\bF_{ij}^1 =& \Big(\frac{1}{n}\sum_{i=1}^n\langle\bbeta_{211}\circ \bbeta_{311},[\cF_1(\cX_i)]_{j..}\rangle, \ldots, \frac{1}{n}\sum_{i=1}^n\langle\bbeta_{2d_n1}\circ \bbeta_{3d_n1},[\cF_{d_n}(\cX_i)]_{j..}\rangle\\
        &\ldots, \frac{1}{n}\sum_{i=1}^n\langle\bbeta_{21R}\circ \bbeta_{31R},[\cF_1(\cX_i)]_{j..}\rangle, \ldots, \frac{1}{n}\sum_{i=1}^n\langle\bbeta_{2d_nR}\circ \bbeta_{3d_nR},[\cF_{d_n}(\cX_i)]_{j..}\rangle\Big),
   \end{split}
   \end{equation*}
   which implies that 
   \begin{equation*}
   \Big\|\frac{2}{n}\sum_{i=1}^n\bF_{ij}^1\Big\|_2= \Big(\sum_{h=1}^{d_n}\sum_{r=1}^R\Big(\frac{2}{n}\sum_{i=1}^n\langle\bbeta_{2hr}\circ\bbeta_{3hr}, [\cF_h(\cX_i)]_{j..}\rangle\Big)^2\Big)^{\tfrac{1}{2}}.
   \end{equation*}
   According to the property of B-spline basis function in Section \ref{subsec:B-spline}, we have
   \begin{equation}\label{eqn:expectation}
   \mathbb E\big\langle\bbeta_{2hr}\circ \bbeta_{3hr}, [\cF_h(\cX_i)]_{j..} \big\rangle\leq d_n^{-1}\sum_{k=1}^p\sum_{l=1}^p\beta_{2hrk}\beta_{3hrl}\leq \frac{C_0s}{d_n}\Big(\sum_{k=1}^p\sum_{l=1}^p\beta_{2hrk}^2\beta_{3hrl}^2\Big)^{\tfrac{1}{2}},
   \end{equation} 
   where the second inequality comes from Cathy-Schwarz inequality and sparsity assumption on $\bb_2,\bb_3$. On the other hand, recall that $\sup_x|\psi_{jklh}(x)|\leq 1$ for all $j,k,l\in[p]$. With the rotation invariance,  the $\phi_2$-Orlicz norm of $\langle\bbeta_{2hr}\circ \bbeta_{3hr}, [\cF_h(\cX_i)]_{j..} \rangle$ can be bounded by $(\sum_{k=1}^p\sum_{l=1}^p\beta_{2hrk}^2\beta_{3hrl}^2)^{\tfrac{1}{2}}$.
  Combining \eqref{eqn:expectation} and Hoeffding-type concentration inequality (See Lemma \ref{lemma:hoffding}), we have with probability at least $1-1/n$,
   \begin{equation}
   \frac{2}{n}\sum_{i=1}^n\langle\bbeta_{2hr}\circ \bbeta_{3hr}, [\cF_h(\cX_i)]_{j..}\rangle\leq 2\Big(\frac{C_0s}{d_n}+\sqrt{\frac{\log (en)}{n}}\Big)\Big(\sum_{k=1}^p\sum_{l=1}^p\beta_{2hrk}^2\beta_{3hrl}^2\Big)^{\tfrac{1}{2}},
   \end{equation}
which implies
   \begin{equation}\label{eqn:init_error}
   \max_{j\in A_1}\Big\|\frac{2}{n}\sum_{i=1}^n\bF_{ij}^1\Big\|_2\leq 2\Big(\frac{C_0s}{d_n}+\sqrt{\frac{\log (en)}{n}}\Big)\Big(\sum_{h=1}^{d_n}\sum_{r=1}^R\sum_{k=1}^p\sum_{l=1}^p\beta_{2hrk}^2\beta_{3hrl}^2\Big)^{\tfrac{1}{2}},
   \end{equation}
   with probability at least $ 1-Rd_ns/n$.
   \item Putting \eqref{eqn:spline_error}-\eqref{eqn:init_error} together, we obtain
   \begin{eqnarray}\label{bound:T1}
   T_1 \leq 2 C_1 s^3d_n^{-\kappa} \Big(\frac{C_0s}{d_n}+\sqrt{\frac{\log (en)}{n}}\Big)\Big(\sum_{h=1}^{d_n}\sum_{r=1}^R\sum_{k=1}^p\sum_{l=1}^p\beta_{2hrk}^2\beta_{3hrl}^2\Big)^{\tfrac{1}{2}}.
   \end{eqnarray}
   with probability at least $1-Rd_ns/n$ for some absolute constant $C_0, C_1$.
\end{enumerate}

\textbf{Step Two: Bounding $T_2$.} Recall that $\bF^{1\top}\bepsilon= (\bF^{1\top}_1\bepsilon, \ldots, \bF^{1\top}_p\bepsilon)^{\top}\in \mathbb R^{pd_n\times 1}$. Then,
\begin{eqnarray*}\label{eqn:T_2}
T_2 &=& \max_{j\in A_1}\Big\|\frac{2}{n}\bF_j^1\bepsilon\Big\|_2 = \max_{j\in A_1}\Big\|\frac{2}{n}\sum_{i=1}^n\bF_{ij}^1\epsilon_i\Big\|_2\\
&\leq&  \frac{1}{\sqrt{n}}\max_{j\in A_1,h\in[d_n], r\in[R]}\sqrt{\frac{d_n}{n}}\sum_{i=1}^n\epsilon_i\langle\bbeta_{2hr}\circ\bbeta_{3hr}, [\cF_h(\cX_i)]_{j..}\rangle\\
&=&\frac{1}{\sqrt{n}}\max_{j\in A_1,h\in[d_n], r\in[R]}\sum_{k\in w(\bb_2)}\sum_{l\in w(\bb_3)}\sqrt{\frac{d_n}{n}}\sum_{i=1}^n\epsilon_i\psi_{jklh}(\cX_i)\beta_{2hrk}\beta_{2hrl}.
\end{eqnarray*}
where the definition of $w(\bx)$ is presented in the beginning of Section \ref{subsec:opt_stat}. From initial value assumption and Condition \ref{con:signal}, we have 
\begin{eqnarray*}
|\beta_{2hrk}-\beta_{2jrk}^*|&\leq& \max_{h,k}|\beta_{2hrk}-\beta_{2hrk}^*|\leq \|\beta_{2hrk}-\beta_{2hrk}^*\|_2\leq \alpha,
\end{eqnarray*}
and thus $\beta_{2hrk}\leq \beta_{2hrk}^*+c^*$. The same result holds for $\beta_{3hk}$. Therefore, by applying Lemma \ref{lemma:max_bound}, we have
\begin{eqnarray}\label{bound:T2}
    T_2 &\leq& \frac{s^2}{\sqrt{n}}(\alpha+c^*)^2\max_{j\in A_1,h\in[d_n], r\in[R]}\sqrt{\frac{d_n}{n}}\sum_{i=1}^n\epsilon_i\psi_{jklh}(\cX_i)\nonumber\\
    &\leq& C_3\sigma\frac{s^2\sqrt{\log(pd_n)}}{\sqrt{n}},
\end{eqnarray}
with probability at least $1-4C_0Rs/n$, where $\sigma$ is the noise level.

\textbf{Step Three: Summary.}
Putting the bounds \eqref{bound:T1} and \eqref{bound:T2} together, we obtain that with probability at least $1-C_0Rd_ns/n$,
\begin{eqnarray*}
&& \Big\|\nabla_1\cL(\bb_1^*, \bb_2, \bb_3)-\nabla_1\tilde{\cL}(\bb_1^*, \bb_2, \bb_3)\Big\|_{\cP^*}\\
&\leq&\Big[C_1 s^3d_n^{-\kappa} \Big(\frac{C_0s}{d_n}+\sqrt{\frac{\log ep}{n}}\Big)\Big]\Big(\sum_{h=1}^{d_n}\sum_{r=1}^R\sum_{k=1}^p\sum_{l=1}^p\beta_{2hrk}^2\beta_{3hrl}^2\Big)^{\tfrac{1}{2}} + C_3\sigma\frac{s^2\sqrt{\log(pd_n)}}{\sqrt{n}}\\
&\leq&\Big[\frac{C_1s^3}{d_n^{\kappa}}\sqrt{\frac{\log ep}{n}}+\frac{C_2s^4}{d_n^{\kappa+1}}\Big]\Big(\sum_{h=1}^{d_n}\sum_{r=1}^R\sum_{k=1}^p\sum_{l=1}^p\beta_{2hrk}^2\beta_{3hrl}^2\Big)^{\tfrac{1}{2}}  + C_3\sigma\frac{s^2\sqrt{\log(pd_n)}}{\sqrt{n}},
\end{eqnarray*}
where $C_1, C_2, C_3$ are some positive constants. According to Condition \ref{con:signal} and  $\bb_2\in \cB_{\alpha, s}(\bb_2^*)$, $\bb_3\in \cB_{\alpha, s}(\bb_3^*)$,
\begin{eqnarray*}
\sum_{h=1}^{d_n}\sum_{r=1}^R\sum_{k=1}^p\sum_{l=1}^p\beta_{2hrk}^2\beta_{3hrl}^2=\sum_{h=1}^{d_n}\Big(\sum_{k=1}^p\beta_{jhk}^2\Big)\Big(\sum_{l=1}^p\beta_{3hl}^2\Big)\leq Rd_n^{1/2}s^2c^{*4}.
\end{eqnarray*}
By setting $C_1=\max\{C_1,C_2,C_2\}$, we have with probability at least $1-C_0Rd_ns/n$,
\begin{eqnarray*}
&&\Big\|\nabla_1\cL(\bb_1^*, \bb_2, \bb_3)-\nabla_1\tilde{\cL}(\bb_1^*, \bb_2, \bb_3)\Big\|_{\cP^*}\\
&\leq& C_1 Rc^{*4}\Big[\frac{s^5}{d_n^{\kappa-1/2}}\sqrt{\frac{\log (en)}{n}}+\frac{s^6}{d_n^{\kappa+1/2}}+\sigma \sqrt{\frac{s^4\log(pd_n)}{n}}\Big].
\end{eqnarray*}
This ends the proof. \hfill $\blacksquare$\\

\section{Supporting Lemmas}

\begin{lemma}[Hoeffding-type inequality]\label{lemma:hoffding}
  Suppose $\{X_i\}_{i=1}^n$ are i.i.d sub-Gaussian random variable with $\|X_i\|_{\phi_2}\leq K$, where $K$ is an absolute constant. For fixed $\ba\in\mathbb R^n$, we have w.p.a $1-\delta$,
  \begin{equation*}
    \Big|\sum_{i=1}^{n}a_iX_i-\mathbb E(\sum_{i=1}^na_iX_i)\Big|\leq C_0 K\|\ba\|_2\sqrt{\log(e/\delta)}.
  \end{equation*}
\end{lemma}

\begin{lemma}[Lemma 2 in \cite{huang2010variable}]\label{lemma:max_bound}
Suppose that Condition \ref{con:nonpara_component}-\ref{con:noise} hold. Let 
\begin{equation*}
    T_{jkl} = \sqrt{\frac{d_n}{n}}\sum_{i=1}^n\psi_{jklh}([\cX_i]_{jkl})\epsilon_i, \ \text{for} \ j\in[p], k\in[p], l\in[p], h\in[d_n],
\end{equation*}
and $T_n = \max_{j,k,l\in[p], h\in[d_n]}|T_{jkl}|$. When $d_n\sqrt{pd_n}/n\rightarrow 0$, we have for some constant $C_1$,
\begin{equation*}
    \mathbb E(T_n) = C_1\sqrt{\log (pd_n)}.
\end{equation*}
\end{lemma}

\end{document}